%% file: iclr2026_conference.tex
\documentclass{article} % For LaTeX2e
\usepackage{iclr2026_conference,times}

\input{math_commands.tex}

\usepackage{hyperref}
\usepackage{url}

\usepackage{amsmath}
\usepackage{cleveref}
\usepackage{algorithm}
\usepackage{algorithmic}
\usepackage{xspace}
\usepackage{soul}
\usepackage{enumitem}
\usepackage{graphicx}
\usepackage{booktabs}
\usepackage{wrapfig}
\usepackage[table]{xcolor}

\newcommand{\modelname}{\textbf{{PRISM}}\xspace}

\title{Seeing Through the Brain: New Insights from Decoding Visual Stimuli with fMRI}

\author{Zheng Huang$^{1}$\thanks{Equal contribution. Correspondence to zheng.huang.gr@dartmouth.edu. },
Enpei Zhang$^{1}$\footnotemark[1],
Weikang Qiu$^{2}$,
Yinghao Cai$^{1}$,
Carl Yang$^{3}$,
Elynn Chen$^{4}$,\\
\textbf{Xiang Zhang$^{5}$,
Rex Ying$^{2}$,
Dawei Zhou$^{6}$,
Yujun Yan$^{1}$}
\\
$^{1}$Dartmouth College \quad
$^{2}$Yale University \quad
$^{3}$Emory University \quad
$^{4}$New York University \\
$^{5}$UNC Charlotte \quad
$^{6}$Virginia Tech
}

\iclrfinalcopy % Uncomment for camera-ready version, but NOT for submission.
\begin{document}

\maketitle
% decoding visual stimuli
\begin{abstract}

% Understanding how the brain represents visual information is a central challenge in neuroscience and machine learning. A promising method is to reconstruct visual stimuli from functional Magnetic Resonance Imaging (fMRI) signals. This involves two steps: (1) transforming fMRI signals to a latent space, and (2) using a generative model to reconstruct the stimuli. The quality of reconstruction therefore depends how well the tranformation between fMRI signals and the latent space, and the capacity of generating the stimuli from the latent space.
% Understanding how the brain encodes visual information is a central challenge in both neuroscience and machine learning. A promising approach to this problem is to reconstruct visual stimuli—essentially images—from functional Magnetic Resonance Imaging (fMRI) signals. This process typically involves two stages: first, transforming fMRI signals into a latent space, and second, reconstructing the images from that space using a generative model. The quality of reconstruction depends on both how similar the latent space is to the structure of neural activity, which makes the transformation from fMRI signals more effective, and how well the generative model can reconstruct images from that space.
Understanding how the brain encodes visual information is a central challenge in neuroscience and machine learning. A promising approach is to reconstruct visual stimuli—essentially images—from functional Magnetic Resonance Imaging (fMRI) signals. This involves two stages: transforming fMRI signals into a latent space and then using a pre-trained generative model to reconstruct images. The reconstruction quality depends on how similar the latent space is to the structure of neural activity and how well the generative model produces images from that space. Yet, it remains unclear which type of latent space best supports this transformation and how it should be organized to represent visual stimuli effectively.

We present two key findings. First, fMRI signals are more similar to the text space of a language model than to either a vision-based space or a joint text–image space. Second, text representations and the generative model should be adapted to capture the compositional nature of visual stimuli, including objects, their detailed attributes, and relationships. Building on these insights, we propose \modelname, a model that {\textbf{P}}rojects fM{\textbf{R}}I s{\textbf{I}}gnals into a \textbf{S}tructured text space as an inter{\textbf{M}}ediate representation for visual stimuli reconstruction. It includes an object-centric diffusion module that generates images by composing individual objects to reduce object detection errors, and an attribute/relationship search module that automatically identifies key attributes and relationships that best align with the neural activity.
Extensive experiments on real-world datasets demonstrate that our framework outperforms existing methods, achieving up to an $6\%$ reduction in perceptual loss. These results highlight the importance of using structured text as an intermediate space to bridge fMRI signals and image reconstruction. Codes are available at \url{https://github.com/GraphmindDartmouth/PRISM}.

\end{abstract}

\input{01introduction}
\input{02preliminary}

\input{03method}

\input{04experiments}

\input{05relatedworks}
\input{06conclusion}

\bibliography{iclr2026_conference}
\bibliographystyle{iclr2026_conference}

\appendix
% \section{Appendix}
% You may include other additional sections here.

\input{10appendix}

\end{document}

%% file: math_commands.tex
\usepackage{amsmath,amsfonts,bm}

\def\eqref#1{equation~\ref{#1}}
\def\1{\bm{1}}

\DeclareMathAlphabet{\mathsfit}{\encodingdefault}{\sfdefault}{m}{sl}
\SetMathAlphabet{\mathsfit}{bold}{\encodingdefault}{\sfdefault}{bx}{n}

%% file: 01introduction.tex
\vspace{-0.2cm}
\section{Introduction}
\vspace{-0.2cm}

Decoding visual stimuli from brain activity provides a unique lens into human perception \citep{naselaris2011encoding, haufe2014interpretation}. A central approach uses fMRI signals—which measure neural activity through blood-oxygen-level-dependent responses—to reconstruct the images perceived by subjects \citep{allen2022massive, chang2019bold5000, luo2023brainscuba}. Recent advances in deep generative models have significantly improved these reconstructions, deepening our understanding of visual representation in the brain \citep{chen2023seeing} and enabling applications in brain-computer interfaces \citep{sitaram2008fmri} and brain-driven content generation \citep{wang2024braindreamer, qiumindllm}.
%\yj{The first and last sentences are good. The middle ones do not connect well. You do not need to mention past work in this paragraph. It would be better that you start with the meaning of decoding human brain, e.g. why do researchers want to decode human brain, what are potential impacts? Among different tools to decode human brains, why using fmri data to generate image is a good way? Why fmri? why image generation?}

% make high level, universal domain, be abstract, concepts, we don't know what the space looks like. we then check if it closes to ML space.
% understanding way is align with cv model processing, if not? close to which kind of.use descriptive language to define. representation space align 

%\yj{The following challenges sound very specific and ill-motivated. It's not clear why we care about these challenges. My suggestions: First talk about general drawbacks of the prior works. In general, say that the prior works focus on the improving generation quality while neglecting the neuroscientific insights. list some papers. Then talk about the three challenges with proper transitions.}
% The success of image-fMRI reconstruction depends on two key factors: the quality of the transformation from fMRI signals to the generative model’s latent space (alignment) and the model’s capacity to generate high-quality images. 
FMRI-to-Image reconstruction involves two stages: mapping fMRI signals into a latent space and then generating images from that space. Its success depends on the similarity between the latent space and neural activity (alignment) and how well the generative model produces high-quality images.
%\yy{please do not use the word synthesize. use generate instead. this is the terminology used in other literature, then follow the literature. same for the remaining usages.}
While recent studies \citep{scotti2023reconstructing, scotti2024mindeye2, mai2024brain} focus on enhancing image quality using advanced generative models \citep{podell2023sdxl, xu2023versatile}, alignment remains underexplored. Prior work often assumes that the latent space should match the modality of the stimuli, i.e., using vision model representations to reconstruct visual stimuli \citep{scotti2023reconstructing, wang2024decoding, xia2024dream}. Some studies incorporate auxiliary semantic information from language models (LMs)~\citep{lin2022mind, quan2024psychometry}, but still rely on vision-based representations as the core latent space. In contrast, we question whether matching the modality of visual stimuli is truly essential for reconstruction.
%\yy{I think first say what other papers did and then throw out the question sounds more logically.}
%Neuroscience shows that regions near the visual cortex border act as convergence zones where visual input transitions into an amodal semantic system \citep{popham2021visual, tang2023semantic}. However, it remains unclear whether this semantic representation aligns more closely with the latent space of vision models or alternative representations. Although some methods incorporate semantic information from language models \citep{lin2022mind, quan2024psychometry}, such information is often auxiliary, with most approaches relying on direct fMRI-to-vision latent mappings \citep{scotti2023reconstructing, wang2024decoding, xia2024dream}. 
In addition, prior work suffers from limited reconstruction quality due to a unified hidden representation that conflates objects and their attributes, often causing object detection errors, e.g., generating a tiger instead of a gray, tiger-striped cat (Section~\ref{app:common_errors} of the Appendix). This reflects a fundamental mismatch with human visual processing, which is object-centric and compositional rather than holistic \citep{marr1980visual, bracci2023understanding}. Overcoming this limitation calls for generative models that explicitly capture the compositional structure of human perception.
%Yet, it remains an open question which object attributes—such as location, geometry, or semantic category—the brain emphasizes during visual representation.
% \yy{Logic a bit wierd here. seems like you are conflicting with what you said before.} \yy{you should first say the current problem at high-level, e.g. poor reconstruction quality. misattributaion error is just an example of this poor reconstruction quality.}
% The fundamental challenge lies in the synthesis process itself: reconstructed images often suffer from generative models misattributing visual properties—for instance, generating a tiger instead of a gray tiger-striped cat (Appendix~\ref{app:common_errors}). This reflects a mismatch with human visual processing, which is object-centric and compositional rather than holistic \citep{marr1980visual, bracci2023understanding}.\yy{why this reflect you should do object-centric and compositional generation? logic gap.} Addressing this gap requires generative models that better capture the object-based structure of human perception. Yet, it remains an open question which object attributes—such as location, geometry, or semantic category—the brain emphasizes during visual representation.\yy{this question also comes from nowhere. you were mentioning that object detection error. why do you suddenly mention object attributes? }

To address these issues, we propose \modelname, a model that {\textbf{P}}rojects fM{\textbf{R}}I s{\textbf{I}}gnals into a \textbf{S}tructured text space as an inter{\textbf{M}}ediate representation for image reconstruction. To identify the most effective intermediate space, we compare fMRI signals with representations from pre-trained vision, language, and vision–language models using established metrics \citep{wang2020finding, murphy2024correcting, keskar2016large}. Unlike prior work assuming vision-based representations are essential, our first finding (\Cref{sec:findings}) shows that fMRI signals align more closely with the text space of an LM, motivating the use of solely text as a bridge for reconstruction. Building on this, our second finding reveals that reconstruction quality improves when the text and the generative model are adapted to capture the compositional and relational nature of visual stimuli—encompassing objects, their attributes, and their relationships. Guided by these insights, we develop two core modules: an object-centric diffusion module that adapts the diffusion model to generate images by composing individual objects, and an attribute/relationship search module that uses a vision–language model (VLM) to automatically identify object attributes and relationships aligned with neural activity, providing structured guidance for reconstruction.
%\yy{you should highlight the difference with the previous works here. something like, unlike prior work that .../contrast to common beliefs that..., our analysis surprisingly show that ...}
% To further enhance reconstruction quality and mitigate object detection errors, we introduce structured texts that encode objects, attributes, and their relationships, inspired by the compositional nature of human perception. Concretely, our framework follows a two-stage process: (1) we adapt the generation process of pre-trained diffusion models to enable object-centric image synthesis, better reflecting the compositional nature of human visual cognition, and (2) we develop a vision–language model–assisted method to automatically discover which object relations are most semantically meaningful and effectively represented in the brain during visual processing. 
Our contributions are summarized as follows:
\begin{itemize}[noitemsep,topsep=0pt]
    \item \textbf{Novel Findings:} To our knowledge, we are the first to show that accurate visual stimuli reconstruction can be achieved without image-based latent representations, with LM text space effectively bridging brain activity and generative models. Furthermore, we find that adapting this text space and the generative model to capture the compositional and relational nature of visual images further improves reconstruction quality.
    %In particular, structured textual descriptions derived from LM space serve as effective guidance for decoding visual stimuli.
    
    %We show that fMRI signals align more closely with the semantic embeddings of language models than with the perceptual features of vision models, suggesting that the brain encodes visual information in an abstract, language-like form.
    \item \textbf{Novel Framework:} 
    % Motivated by our empirical findings,
    % %\yy{if you use this word, make sure you mark it in the previous text} 
    % we propose a new fMRI-to-Image reconstruction framework, which enables object-centric reconstruction by adapting diffusion models and automatically discovers brain-aligned object relations and relations using vision–language models.
    Motivated by our empirical findings, we introduce a new fMRI-to-image reconstruction framework that adapts diffusion models for object-centric generation and leverages VLMs to automatically identify brain-aligned object attributes and relationships that can optimally guide the reconstruction.
    \item \textbf{Comprehensive Experiments:} Extensive evaluations on real-world fMRI datasets demonstrate that our method achieves up to an $6\%$ reduction in perceptual loss compared to state-of-the-art models, highlighting the effectiveness of our framework.
\end{itemize}

% to address misattribute, constrain text-> text object -> object span multi details (optimal attributes)-> strctual text
% strctual text -> t5 . combined with learned t5 + object diffusion -> recon

%% file: 02preliminary.tex
\vspace{-0.1cm}
\section{Preliminary}
\label{sec:prelim}
\vspace{-0.2cm}

% diffusion Diffusion 
% clip img encoder, text enc
% text embedding: 
% ENC_{img}(I), ENC_{text}(T)
% remove A

% \textbf{Functional Magnetic Resonance Imaging (fMRI). } 
% Each raw fMRI sample is a 4D volume (space and time), averaged over time to produce a 3D brain volume per stimulus. This is then flattened into a 1D vector to work with machine learning models, with each element representing a single voxel’s neural activation~\citep{allen2022massive}.

\textbf{Notations. } 
In our work, we denote the set of fMRI samples collected during image viewing as $\mathcal{X}$. Each sample is a preprocessed 1D vector $\mathbf{x}_i \in \mathbb{R}^v$, capturing neural activity across $v$ voxels selected from brain regions \citep{scotti2023reconstructing}. The dataset is split into training and test subsets, with superscripts indicating the split. For example, we denote the training set with $N$ samples as: $\mathcal{X}^{\text{train}} = \{\mathbf{x}_1, \ldots, \mathbf{x}_N\}$. The corresponding image stimulus for the $i$-th sample is denoted as: $\mathbf{Y}_i \in \mathbb{R}^{H \times W \times 3}$, which contains $m$ objects. 

$\textbf{Problem Setup. }$ Our goal is to reconstruct the visual images that subjects viewed during fMRI recording. Formally, we seek to learn a reconstruction function $\mathcal{F}: \mathbb{R}^v \rightarrow \mathbb{R}^{H \times W \times 3}$ that maps each fMRI sample $\mathbf{x}_i$ to its corresponding image stimulus $\mathbf{Y}_i$.
% In our framework, we instantiate $\mathcal{F}$ as a composition of three components: an multi-layer perceptron (MLP) that maps $X_i$ into a language embedding space, a language model (LM) that generates structured descriptions, and a diffusion model that reconstructs the image:
% $
% \mathcal{F}(X_i) = \text{Diffusion}(\text{LM}(\text{MLP}(X_i))).
% $ The objective of the fmri-image reconstuction is to minimize the perceptual loss between $\hat{I}_i$ and $I_i$.

% diffusion pipeline, basic diffusion, various diffusion
$\textbf{Diffusion Model. }$ 
Diffusion models~\citep{rombach2022high, zhang2023adding} are a class of generative models that synthesize data by learning to reverse a multi-step noising process. Starting from Gaussian noise, they iteratively denoise a latent variable over a fixed number of timesteps using a denoising network—typically a U-Net—conditioned on the current timestep $t$. This process gradually produces samples resembling the training distribution.
 % \yj{lack citation for this conditional generation}
To incorporate external inputs such as text, the denoising network can take a conditioning input $\mathbf{C}$, usually text embeddings from a pre-trained encoder~\citep{zhang2023adding}. Conditioning is implemented via cross-attention mechanisms within the U-Net architecture~\citep{williams2023unified}, enabling the integration of textual information. In these layers, the latent representations $\mathbf{H}_t \in \mathbb{R}^{h \times w \times d}$ at time $t$ serve as queries, and $\mathbf{C} \in \mathbb{R}^{d_t \times d'}$ serves as both keys and values~\citep{williams2023unified, yang2024mastering}: 
%\yj{check formula and meaning}
\begin{equation*}
\vspace{-0.1cm}
\text{CrossAttention}(\mathbf{H}_t, \mathbf{C}) = \text{softmax} \left( \frac{ \phi(\mathbf{H}_t) \cdot \mathbf{W}_Q \cdot (\varphi(\mathbf{C}) \cdot \mathbf{W}_{K})^\top}{\sqrt{d_k}} \right) \varphi(\mathbf{C}) \cdot \mathbf{W}_{V},
\end{equation*}
where $\mathbf{W}_Q\in \mathbb{R}^{d \times d_k} $ ,  $\mathbf{W}_K \in \mathbb{R}^{d^\prime \times d_k} $, $\mathbf{W}_V \in \mathbb{R}^{d^\prime \times d}$ are projection matrices; and $ \phi(\cdot)$ and $ \varphi(\cdot)$ are learned transformations. Further details are available in Section~\ref{app:diffusion} of the Appendix.

%% file: 03method.tex
\section{Method}
\label{sec:method}
% based on design, (callback 3.1) 1. text bridge
% subsubsection{1} have a brief intro, why align(difference between mindeye, solely text is enough), why use text xxx alignment
% subsection{2} the corresponding design: fmri text img. use a var as text, mlp, supervised by text

% compositional gene, existing insight: xxx, based on it we design
% formatting: 
%   then optimal object level description
% cross att

% prompt opt: auto text labeling, different prompt return different description. use llm to refine the keywords. the prompt can better label the images. then the description can be used to better recon

\vspace{-0.4cm}
\begin{figure}[t]
    \centering
    \includegraphics[width=0.98\linewidth]{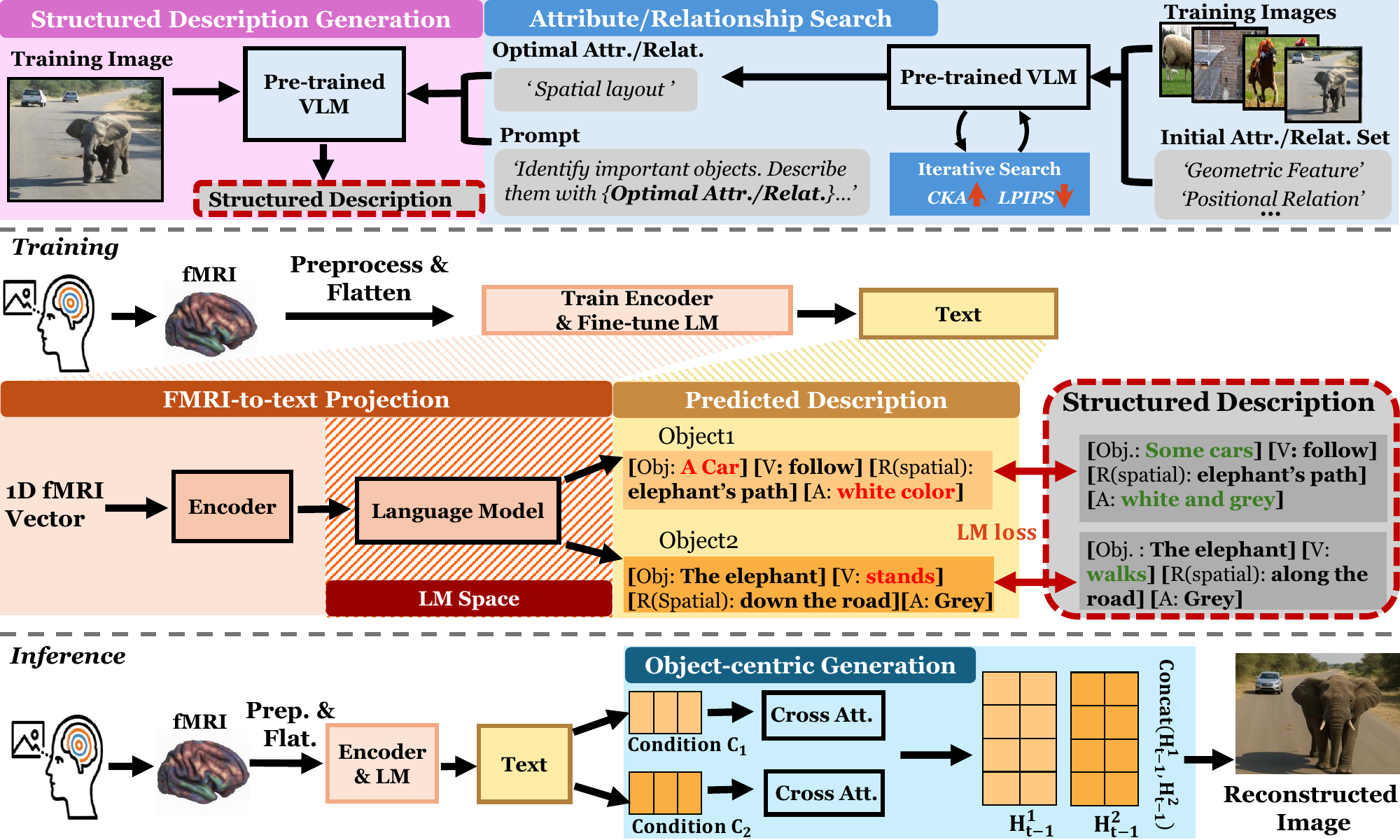}
    \caption{\small Framework Overview: \modelname generates structured text descriptions for each training image using a VLM to iteratively extract brain-aligned object attributes and relationships. These descriptions capture the image’s compositional and relational content and serve as supervision to train an encoder and fine-tune a language model to map fMRI signals into the text space. During inference, the model predicts descriptions from fMRI signals, which then guide a pre-trained diffusion model for object-centric image reconstruction.}
    \label{fig:framework}
    \vspace{-0.25cm}
\end{figure}

In this section, we present our framework, \modelname, for fMRI-to-image reconstruction (Figure~\ref{fig:framework}). We first show that fMRI signals align most strongly with the text space of LMs, compared to the hidden spaces of vision or vision–language models, under established metrics (\Cref{sec:findings}). This finding motivates our choice of using pure text as the latent space.
During training (\Cref{sec: text_bridge}), we annotate each training image with structured text descriptions that are object-centric, compositional, and relational. To generate these descriptions, we introduce an attribute/relationship search module (\Cref{sec:train_desc_gen}), which learns optimal prompts to guide the VLM in automatically identifying the key attributes and relationships most aligned with both the fMRI signals and images. These structured descriptions are then used to train an encoder and fine-tune the LM, mapping fMRI signals into the LM text space (\Cref{sec:encoder_training}).
At inference time (\Cref{sec:inference}), the predicted structured descriptions guide an adapted diffusion model to generate object-centric images directly from fMRI signals.

\subsection{Text as the Latent Space}
\label{sec:findings}
 We question whether using vision representations as the latent space is truly essential for reconstructing visual stimuli. In this section, we investigate the alignment between different model spaces and fMRI signals using various measures. 

\textbf{Measuring the alignment between model spaces and fMRI signals.}
We examine three representation spaces: (1) the text space of language models, (2) the joint text-image space of vision-language models, and (3) the latent space of vision models. For (2) and (3), image embeddings are extracted directly from the respective models. For (1), we use text embeddings from image captions to represent the stimuli. We extract embeddings by feeding either text or images into different models: T5 and LLaMA3
%\textcolor{blue}{Qwen3~\citep{yang2025qwen3}  and RoBERTa}~\citep{liu2019roberta} 
for text embeddings, 
%\textcolor{blue}{SD3}~\citep{esser2024scaling} 
LDM~\citep{rombach2022high} and ResNet50~\citep{he2016deep} for image embeddings, and CLIP for both modalities.

Alignment is assessed using three metrics: Centered Kernel Alignment (CKA) \citep{murphy2024correcting}, Canonical Correlation Analysis (CCA)~\citep{wang2020finding}, and Generalization Gap~\citep{keskar2016large}. CKA and CCA are widely used to quantify similarity between representation spaces \citep{kriegeskorte2008representational, wang2020finding}. Generalization Gap reflects learnability by measuring the train-test loss difference when mapping fMRI signals to a target space using an MLP. Good alignment yields higher CKA and CCA values and a lower Generalization Gap.

Let \( \mathbf{X} = \text{Concat}(\mathbf{x}_1, \ldots, \mathbf{x}_N) \) denote concatenated fMRI samples and \( \mathbf{K} = \text{Concat}(\mathbf{k}_1, \ldots, \mathbf{k}_N) \) the corresponding latent representations. With \( \mathcal{K}(\cdot) \) as a kernel function, the empirical Hilbert-Schmidt Independence Criterion (HSIC) is:
$
\text{HSIC}(\mathbf{X}, \mathbf{K}) = \frac{1}{(N - 1)^2} \operatorname{tr} \left( \mathcal{K}(\mathbf{X}) \cdot \mathcal{K}(\mathbf{K}) \right),
$
where \( \operatorname{tr}(\cdot) \) denotes the trace operator. The CKA is the normalized form:
$
\text{CKA}(\mathbf{X}, \mathbf{K}) = \frac{\text{HSIC}(\mathbf{X}, \mathbf{K})}{\sqrt{\text{HSIC}(\mathbf{X}, \mathbf{X}) \cdot \text{HSIC}(\mathbf{K}, \mathbf{K})}}.
$
We adopt a Gaussian radial basis function (RBF) kernel for \( \mathcal{K} \) \citep{alvarez2022gaussian, cortes2012algorithms}. CCA identifies linear projections $\mathbf{u} = \mathbf{p}_1^\top \mathbf{X}$ and $\mathbf{v} = \mathbf{p}_2^\top \mathbf{K}$ that maximize their correlation. The first mode captures the dominant shared axis \citep{wang2020finding}, with the canonical correlation coefficient: $\rho = \text{corr}(\mathbf{u}, \mathbf{v}) = \text{corr}(\mathbf{p}_1^\top \mathbf{X}, \mathbf{p}_2^\top \mathbf{K}),$ reflecting the strongest linear alignment between brain activity and the model space.
\begin{wraptable}{r}{0.45\textwidth}  
\vspace{-0.2cm}
\centering
    \caption{\small Alignment results between model representations and fMRI data, evaluated using CKA, Generalization Gap, and CCA. The best result is highlighted in \textcolor{red}{red}. $\uparrow$ denotes higher is better; $\downarrow$ denotes lower is better.}
    \vspace{0.2cm}
    \resizebox{0.9\linewidth}{!}{

    \begin{tabular}{lccc}
    \toprule
               & \textbf{CKA} $\uparrow$   & \textbf{Generalization Gap}  $\downarrow$ & \textbf{CCA} $\uparrow$\\
    \midrule
    \textbf{T5}    & \textcolor{red}{0.5580}  & \textcolor{red}{0.1132}             &  \textcolor{red}{0.8344}   \\
    \textbf{Llama3} & 0.5442 & 0.2216             &  0.8022   \\
    \textbf{Clip text}  & 0.5177 & 0.4532             &   0.7599  \\
    \textbf{Clip img}   & 0.3668 & 0.4860                &   0.7573  \\
    \textbf{LDM}        & 0.1957 & 1.2520              & 0.7215    \\
    \textbf{Resnet50}     & 0.1822 & 1.9800               &  0.6746  \\
    \bottomrule

    \end{tabular}
    \label{tb_similarity}

    }
\vspace{-0.4cm}
\end{wraptable}

\textbf{FMRI aligns better with the embedding space of language models.} Our results (Table~\ref{tb_similarity}) show that the text space of language models aligns best with fMRI data, outperforming both vision–language and vision-only models across all metrics. Surprisingly, vision–language models, despite integrating both modalities, underperform compared to pure language models. We hypothesize that this is because humans focus more on the meaning of an image rather than pixel-level details~\citep{naselaris2009bayesian, du2022fmri}. Unlike prior work \citep{scotti2023reconstructing, wang2024decoding, xia2024dream, lin2022mind} that primarily relies on vision representations, our findings motivate using pure text as the latent space.

\subsection{Training of \modelname}
\label{sec: text_bridge}
% Building on our findings in \Cref{sec: finding_text}, we adopt the language model’s text space as a semantic bridge for fMRI-to-image generation. Our approach first maps fMRI signals into a language model’s embedding space and then generates textual descriptions that are subsequently linked to the vision model.
% In this section, we describe how PERM is trained to generate structured, object-level descriptions that capture the compositional nature of human visual perception.
In this section, we describe the training process of \modelname, which consists of automatic structured description generation for training images and encoder training.

\subsubsection{Automatic Description Generation}
\label{sec:train_desc_gen}
We design structured text descriptions as supervision for our framework. To capture the compositional and relational nature of human vision, these descriptions should explicitly distinguish between different objects and their relationships. Generating such descriptions with a VLM relies on carefully crafted prompts that specify the desired attributes and relationships, since many of them are not directly reflected in brain activity. To address this issue, we propose a VLM-assisted approach that automatically learns the most relevant attributes and relationships in an image based on the training data, ensuring they are both meaningful and brain-aligned.

%\yy{needs rewriting, notation issues, also, not only relationship, but also attribute, you can say a is a keyword, equation 1 useless}Given the $i$-th fMRI sample $\mathbf{x}_i \in \mathbb{R}^v$ and its corresponding stmimuli image's caption $D_i$\yy{not given}, 

We first show how structured descriptions can be generated from a VLM given a learned keyword $a$, and then present our approach for learning the optimal keyword. Given an image $\mathbf{Y}_i$ and a learned keyword $a$, we construct a prompt $\mathcal{P}(a)$ to guide the VLM in describing the most important objects in $\mathbf{Y}_i$ based on $a$. Formally, the VLM receives the image and the prompt as input and outputs a structured description $D_i^a$:
%(referred to as $D_i$ in \Cref{sec: text_bridge}):
\begin{equation}
    D_i^a = \text{VLM}(\mathbf{Y}_i, \mathcal{P}(a)).
%\vspace{-0.4cm}
\end{equation}

\vspace{-0.3cm}
The structured description is a list of $m$ object-level tuples along with background information:
% \vspace{-0.1cm}

\vspace{-0.3cm}
\begin{equation}
D_i^a = \left[ (o_1: d_1: \text{loc}_1), \; (o_2: d_2 : \text{loc}_2), \; \dots, \;( o_m: d_m: \text{loc}_m), bg_i \right].
% \vspace{-0.2cm}
\label{eq:structure_des}
\end{equation}

Each $o_j$ is an object in the image, $d_j$ is its description containing attributes and relationships with other objects conditioned on keyword $a$, and $\text{loc}_j$ denotes its location (selected from a predefined set). The term $bg_i$ represents the background information of image $\mathbf{Y}_i$. 
% This procedure ensures that the generated description is semantically structured, object-centric, and tailored to the chosen attribute. 
To ensure meaningful generation, we further augment each relation description $ d_j$ with a structured header encoding its semantic roles, following the PropBank annotation format~\citep{marquez2008semantic, he2017deep, ross2021tailor, palmer2005proposition}. \textcolor{black}{A case study on reconstruction with object-level description is provided in Section~\ref{app:case_obj_level} of the Appendix}.
%These headers annotate each sentence with argument structures,
%(e.g., [Obj], [V], [Attribute (A)], [Relationship (R)])
%as shown in Figure~\ref{fig:framework}.

% Object-level descriptions of training images are conditioned on a keyword $a$, which is expected to capture the attributes and relationships most relevant to brain activity. Since the choice of $a$ critically determines the semantic quality and neural alignment of the resulting descriptions, we cast keyword identification as a prompt optimization problem.
The choice of keyword $a$ strongly influences the attributes and relationships captured in the object descriptions, which in turn affects the quality of mapping fMRI signals to the text space. Ideally, these descriptions should capture the most important information shared between the fMRI signals and the stimulus images. To avoid manually selecting the keyword, we frame its discovery as a prompt optimization problem and introduce our attribute/relationship search module.
%Our objective is to identify the optimal keyword $a$ that produces high-quality descriptions $D_i^a$ for reconstructing the original image $\mathbf{Y}_i$.

%To investigate which attribute $a$ used by the VLM can generate the most effective descriptions $D_i^a$ to reconstruct the original image $I_i$, we formulate this challenge as a prompt optimization problem. 

Concretely, given a set of training images $\mathcal{Y}^{\text{train}} = \{\mathbf{Y}_1, \ldots, \mathbf{Y}_N\}$ and the corresponding fMRI signals $\mathcal{X}^{\text{train}} = \{\mathbf{x}_1, \ldots, \mathbf{x}_N\}$, we define the following optimization problem to find the optimal $a$ in the prompt ($\mathcal{P}(a)$) for the VLM:
\vspace{-0.2cm}
\begin{align}
\max_{a} \quad & \sum_{i=1}^N\mathcal{S}\left(\mathbf{Y}_i,\text{Diff}(\text{VLM}(\mathbf{Y}_i,\mathcal{P}(a)))\right) \label{eq:opt_obj} \\
\text{s.t.} \quad & \text{CKA}\left(\mathbf{X}, \mathbf{K}^a\right) > \beta; \nonumber\\
& \mathbf{X}=\text{Concat}(\mathbf{x}_1,\ldots,\mathbf{x}_N);~\mathbf{K}^a=\text{Concat}(\mathbf{k}_1^a,\ldots,\mathbf{k}_N^a); \nonumber\\
& \mathbf{k}_i^a = \text{LM}_{\text{ENC}}(\text{VLM}(\mathbf{Y}_i,\mathcal{P}(a)))~\text{for } i=1,\ldots,N \nonumber;
\end{align}
where $\mathcal{S}(\cdot, \cdot)$ denotes the similarity score between two images (e.g. negative perceptual loss); $\text{Diff}$ is a pre-trained diffusion model that generates images from captions produced by the VLM; $\text{Concat}$ indicates the concatenation operation across all training samples; and $\text{LM}_{\text{ENC}}$ is a pre-trained language model to encode captions generated by the VLM. The constraint enforces that the CKA similarity between the fMRI data $\mathbf{X}$ and the caption embeddings $\mathbf{K}^a$ generated using keyword $a$ exceeds a threshold $\beta$, ensuring strong alignment between the fMRI and text spaces. The objective ensures that descriptions derived from the optimal keyword support accurate reconstruction.

To optimize the keyword $a$ in~\Cref{eq:opt_obj}, we guide the search along semantic links: keywords with similar meanings tend to yield comparable reconstructions, so generating new keywords based on the semantic relationships of top-performing candidates helps uncover more effective prompt expressions. In the search, we utilize an LLM as a keyword generator and iteratively search for improved keywords in a step-by-step manner. We begin by initializing a keywords set $\mathcal{A}$ with a collection of frequently-used attribute and relationship keywords identified in prior works~\citep{johnson2015image,lu2016visual,krishna2017visual}. We expand $\mathcal{A}$ through an $\varepsilon$-greedy search strategy: at each search step, the keyword generator proposes new candidate keywords based on either the top-performing keywords in $\mathcal{A}$ with probability $1{-}\varepsilon$, or randomly selected keywords from $\mathcal{A}$ with probability $\varepsilon$. Only candidates that exceed the similarity threshold are added to $\mathcal{A}$. This balances refinement of effective keywords and exploration of diverse novel keywords. See Section~\ref{app:attr_opt} in the Appendix for the detailed algorithm and search results.

\subsubsection{Encoder Training}
\label{sec:encoder_training}
We design an encoder to map fMRI signals into the latent space of the language model, using structured and object-centric descriptions as supervision. Specifically, each object's information is independently encoded using an MLP. The resulting representations are concatenated and passed to the language model to generate estimated structured descriptions  $\hat{D}_i^a$, which can be expressed as:

% map the fMRI signals $x_i$ to each object's latent representation $\mathbf{f}_j$, concatnate these representations and feed to a 

% As a result, we are able to map fMRI signal $x_i$ into multiple objects' embeddings:
\vspace{-0.2cm}
% $$
% \mathbf{f}_1, \mathbf{f}_2, \dots, \mathbf{f}_m = \text{MLP}_1(\mathbf{x}_{i}),
% $$
% where $\mathbf{f}_j$ is the object's representation.  
% Specifically, the structured description can be predicted by:
\vspace{-0.3cm}
\begin{equation}
\label{eq:d_object}
\begin{split}
\mathbf{f}_j &= \text{MLP}_j(\mathbf{x}_{i}), \; j= 1, \cdots, m\\
    \hat{D}_i^a &= \text{LM}(\text{MLP}_g(\text{Concat}(\mathbf{f}_1, \dots, \mathbf{f}_m))),
\end{split}
\vspace{-0.2cm}
\end{equation}
% Consequently, the language model generates $m$ structured descriptions, each corresponding to an object in the image, along with a background description. The language model is thus optimized using a sum over $m$ language modeling losses~\citep{chang2024survey, gunel2020supervised}:
The language model is fine-tuned using a loss over all $m$ object descriptions~\citep{chang2024survey, gunel2020supervised}:
\vspace{-0.4cm}
\begin{equation}
\mathcal{L}_{\text{LM}} = - \sum_{j=1}^m \sum_{t'=1}^{T} \log p(y_{t'} \mid y_{<t'}, \mathbf{f}_j),
\end{equation}
where $ y_{t'}$ denotes the $t'$-th token in the structured description. This training strategy enables fine-grained alignment between fMRI signals and structured textual descriptions. We first train the MLPs independently for a fixed number of epochs, then jointly fine-tune the language model and MLPs to maximize overall reconstruction performance.

\subsection{\modelname Inference: Object-Centric Image Generation}
\label{sec:inference}
The inference process of \modelname has two steps: (1) generate structured descriptions $\hat{D}_i^a$ from fMRI signals using the trained encoder and language model, and (2) reconstruct the image by composing objects conditioned on these descriptions with a pre-trained diffusion model: $\hat{\mathbf{Y}}_i =\text{Diff}(\hat{D}_i^a)$. 
% \begin{equation}
% \label{eq:d}
% \begin{split}
%   %\hat{D}_i^a &= \text{LM} (\text{Encoder}(\mathbf{x}_{i})),\\
% \hat{\mathbf{Y}}_i &=\text{Diff}(\hat{D}_i^a).  
% \end{split}
% \vspace{-0.9cm}
% \end{equation}
% \vspace{-0.2cm}

To reflect the brain's compositional understanding of visual scenes, during inference, we adapt a pre-trained diffusion model to perform compositional image generation, inspired by~\citep{yang2024mastering}. Specifically,
given an image $\mathbf{Y}_i$ and its predicted structured description $\hat{D}_i^a$ (from~\Cref{eq:d_object}), we extract a set of predicted objects $\{o_j\}_{j=1}^m$ and a background description $\hat{bg}_i$. These are combined into a global context prompt $\hat{p}_0$ and embedded into a conditioning matrix $\mathbf{C}_0$. Similarly, each object description $\hat{d}_j$ is embedded into a corresponding conditioning matrix $\mathbf{C}_j$. These conditioning matrices guide the denoising process via cross-attention (\Cref{sec:prelim}) from time $t$ to 0. At each step, the hidden representation of object $j$ (or the global context when $j = 0$) at time $t-1$ is computed as: $\mathbf{H}_{t-1}^j = \text{CrossAttention}(\mathbf{H}_t, \mathbf{C}_j),$ where $\mathbf{H}_t$ is the hidden representation of the full image at time $t$. We then resize and concatenate the object representations according to their predicted locations $\hat{\text{loc}}_j$:
\begin{equation}
    \mathbf{H}_{t-1}^{\text{cat}}=\Psi(\{\mathbf{H}_{t-1}^j, \hat{\text{loc}}_j\}_{j=1}^m),
\end{equation}
where $\Psi(\cdot)$ denotes the resizing and spatially-aware concatenation operation. Thus, the image generation model encodes each object independently from its description and then spatially concatenates their hidden representations according to the predicted locations.
%\yy{one sentence to explain what is means as questioned in the rebuttal.}

To ensure smooth region boundaries and seamless fusion between objects and background, we compute a weighted sum of the global context latent and the object latents:
\begin{equation}
    \mathbf{H}_{t-1} = \beta \cdot \mathbf{H}_{t-1}^{\text{cat}} + (1 - \beta) \cdot \mathbf{H}_{t-1}^{0},
\end{equation}

where $\beta$ is a hyperparameter that controls the blending ratio. This process is repeated across denoising steps, enabling structured, object-aware generation aligned with the brain's visual understanding.

%% file: 04experiments.tex
\section{Experiments}
\label{sec:exp}

In this section, we conduct extensive experiments to evaluate \modelname, guided by the following questions: 
%(\textbf{RQ1}) Does the latent space of image models align well with fMRI signals? If not, which representation space shows stronger correspondence? 
(\textbf{RQ1}) How well does our framework \modelname perform on the image reconstruction task? (\textbf{RQ2}) How do different choices of latent space influence the reconstruction quality? (\textbf{RQ3}) What is the contribution of each component in our framework to the overall reconstruction performance?

\subsection{Experimental Setup}
We conduct experiments on three datasets: NSD \citep{allen2022massive}, BOLD5000 \citep{chang2019bold5000}, and GOD \citep{horikawa2017generic}. Detailed descriptions of the datasets are provided in Section~\ref{app:dataset} of the Appendix. Each method is evaluated by comparing the reconstructed images to the ground truth using five metrics: PixCorr and SSIM for pixel- and structure-level similarity \citep{wang2004image}, LPIPS \citep{zhang2018unreasonable} for human perceptual similarity, and CLIP and Inception V3 two-way identification \citep{scotti2024mindeye2} for semantic consistency based on pretrained model representations.
%Pixelwise Correlation (PixCorr), Structural Similarity Index (SSIM) \citep{wang2004image}, Learned Perceptual Image Patch Similarity (LPIPS) \citep{zhang2018unreasonable}, which reflects human perceptual similarity, CLIP two-way identification (CLIP) \cite{scotti2024mindeye2} and Inception V3 two-way identification (Inception V3) \cite{scotti2024mindeye2}.
We compare our method against the following baselines: Takagi \& Nishimoto \citep{takagi2023high} (Takagi for short), Mindvis \citep{chen2023seeing}, Mindeye \citep{scotti2023reconstructing}, \textcolor{black}{MindBridge~\citep{wang2024mindbridge}, NeuralDiffuser~\citep{li2025neuraldiffuser}}, and Mindeye2 \citep{scotti2024mindeye2}. To ensure a fair comparison, we use the same generative model, Stable Diffusion 2.1~\citep{pernias2023wurstchen, rombach2022high}, for all methods. We additionally present results for \modelname and Mindeye2 (ranked second-best) with the newer SDXL backbone~\citep{podell2023sdxl}. \textcolor{black}{We incorporate a negative prompt in our model—a textual constraint that guides the diffusion model to avoid generating undesired visual artifacts, such as distorted object shapes or background clutter. For fairness, we apply the same negative prompt when evaluating all baselines, which leads to performance gains in some cases. } More details about the baselines and training are provided in Sections~\ref{app:baseline} and~\ref{app:training} of the Appendix.

\subsection{Effectiveness of \modelname }

\begin{table}[t] 
\centering
\caption{\small Comparison of our framework with state-of-the-art methods on three datasets. All methods use Stable Diffusion 2.1 as the backbone unless otherwise specified (+SDXL). Results are reported using PixCorr, SSIM, LPIPS, CLIP and Inception V3 metrics. The best result using the same backbone in each column is highlighted in \textcolor{red}{red}. $\uparrow$ indicates higher is better and $\downarrow$ indicates lower is better. }
\resizebox{0.9\textwidth}{!}{
\begin{tabular}{lccccc}
\toprule
          
          \multicolumn{1}{l}{\textbf{\small NSD}} & \textbf{PixCorr} $\uparrow$ & \textbf{SSIM} $\uparrow$  & \textbf{LPIPS} $\downarrow$  & \textbf{CLIP} $\uparrow$  & \textbf{Inception V3} $\uparrow$  \\

\midrule
\textbf{\modelname}      & \textcolor{red}{0.3404$_{\pm 0.05}$}  & \textcolor{red}{0.4640$_{\pm 0.02}$}  & \textcolor{red}{0.5963$_{\pm 0.02}$} &     \textcolor{red}{0.9467$_{\pm 0.03}$}   & \textcolor{red}{0.9516$_{\pm 0.03}$}  \\
\textbf{Takagi} & 0.2100$_{\pm 0.01}$    & 0.3880$_{\pm 0.04}$ & 0.7665$_{\pm 0.04}$ & 0.8811$_{\pm 0.06}$  &  0.9086$_{\pm 0.07}$     \\
\textbf{Mindvis}   & 0.2736$_{\pm 0.06}$  & 0.3868$_{\pm 0.06}$ & 0.6789$_{\pm 0.02}$ & 0.9000 $_{\pm 0.05}$ & 0.9135 $_{\pm 0.05}$ \\
\textbf{Mindeye1}  & 0.3114$_{\pm 0.05}$  & 0.3868$_{\pm 0.06}$ & 0.6501$_{\pm 0.03}$ & 0.9121 $_{\pm 0.04}$ & 0.9198$_{\pm 0.03}$
\\
\textcolor{black}{\textbf{MindBridge}}   & \textcolor{black}{0.1802$_{\pm 0.03}$} & \textcolor{black}{0.2823$_{\pm 0.02}$} & \textcolor{black}{0.6977$_{\pm 0.03}$} & \textcolor{black}{0.9427$_{\pm 0.02}$} & 0.9242$_{\pm 0.03}$
\\
\textcolor{black}{\textbf{NeuralDiffuser}} & \textcolor{black}{0.3011$_{\pm 0.05}$} & \textcolor{black}{0.3348$_{\pm 0.03}$} & \textcolor{black}{0.6522$_{\pm 0.04}$} & \textcolor{black}{0.9409$_{\pm 0.01}$} & \textcolor{black}{0.9487$_{\pm 0.02}$}
\\
\textbf{Mindeye2}  & 0.3160$_{\pm 0.04}$   & 0.4447$_{\pm 0.02}$ & 0.6338$_{\pm 0.04}$ & 0.9201$_{\pm 0.03}$ & 0.9308$_{\pm 0.03}$  \\ 
          \midrule

\textbf{\modelname}+SDXL & \textcolor{red}{0.3645$_{\pm 0.02}$}   & \textcolor{red}{0.4983$_{\pm 0.04}$} & \textcolor{red}{0.5563$_{\pm 0.02}$} & \textcolor{red}{0.9600$_{\pm 0.01}$} & \textcolor{red}{0.9765$_{\pm 0.01}$}  \\

\textbf{Mde2}+SDXL  & 0.3471$_{\pm 0.04}$   & 0.4425$_{\pm 0.04}$ & 0.6002$_{\pm 0.01}$ & 0.9599$_{\pm 0.02}$ & 0.9602$_{\pm 0.01}$  \\

\midrule

\multicolumn{5}{l}{\textbf{\small BOLD5000}} &   \\

\midrule

\textbf{\modelname} &\textcolor{red}{0.2315$_{\pm 0.01}$}   & \textcolor{red}{0.5341$_{\pm 0.02}$}   & \textcolor{red}{0.6198$_{\pm 0.02}$}  & \textcolor{red}{0.7720$_{\pm 0.03}$} & \textcolor{red}{0.6601$_{\pm 0.07}$} \\

\textbf{Takagi} &0.1815$_{\pm 0.03}$   & 0.4418$_{\pm 0.06}$ & 0.7558$_{\pm 0.06}$ & 0.6990$_{\pm 0.04}$ & 0.5667$_{\pm 0.01}$ \\

\textbf{Mindvis} & 0.2122 $_{\pm 0.05}$  & 0.4944 $_{\pm 0.04}$& 0.6463 $_{\pm 0.05}$
& 0.7720$_{\pm 0.04}$ & 0.5701 $_{\pm 0.08}$ \\

\textbf{Mindeye1} & 0.1942$_{\pm 0.01}$    & 0.4838$_{\pm 0.03}$  & 0.6913$_{\pm 0.04}$  & 0.7288 $_{\pm 0.03}$ &0.6222 $_{\pm 0.07}$ \\
\textcolor{black}{
\textbf{MindBridge} }    & \textcolor{black}{0.1522$_{\pm 0.04}$} & \textcolor{black}{0.3005$_{\pm 0.01}$} & \textcolor{black}{0.6535$_{\pm 0.03}$} & \textcolor{black}{0.7431$_{\pm 0.01}$} & \textcolor{black}{0.6200$_{\pm 0.04}$}
\\
\textcolor{black}{\textbf{NeuralDiffuser}} & \textcolor{black}{0.2036$_{\pm 0.04}$} & \textcolor{black}{0.4005$_{\pm 0.02}$} & \textcolor{black}{0.6899$_{\pm 0.01}$} & \textcolor{black}{0.7609$_{\pm 0.01}$} & \textcolor{black}{0.6522$_{\pm 0.03}$}
\\

\textbf{Mindeye2} & 0.2265$_{\pm 0.02}$  & 0.5164$_{\pm 0.02}$  & 0.6416$_{\pm 0.03}$  & 0.7600$_{\pm 0.04}$ & 0.6428$_{\pm 0.03}$ \\

\midrule
\textbf{\modelname}+SDXL &\textcolor{red}{0.2442 $_{\pm 0.02}$}  & \textcolor{red}{0.5600 $_{\pm 0.03}$} & \textcolor{red}{0.5909 $_{\pm 0.04}$}  & \textcolor{red}{0.7881$_{\pm 0.04}$} & \textcolor{red}{0.6881$_{\pm 0.02}$}\\

\textbf{Mde2}+SDXL &0.2310 $_{\pm 0.04}$  & 0.5185 $_{\pm 0.02}$ & 0.6186 $_{\pm 0.02}$  & 0.7503$_{\pm 0.04}$ & 0.6556$_{\pm 0.02}$ \\
\midrule

\multicolumn{5}{l}{\textbf{\small GOD}} &   \\
\midrule

\textbf{\modelname} & \textcolor{red}{0.2571 $_{\pm 0.01}$}   & \textcolor{red}{0.5200 $_{\pm 0.02}$} & \textcolor{red}{0.6213 $_{\pm 0.01}$} & \textcolor{red}{0.8567 $_{\pm 0.05}$} & \textcolor{red}{0.8428 $_{\pm 0.06}$} \\

\textbf{Takagi} & 0.2322 $_{\pm 0.05}$  & 0.4944 $_{\pm 0.04}$& 0.6463 $_{\pm 0.05}$ & 0.7232 $_{\pm 0.01}$ & 0.7556 $_{\pm 0.02}$ \\

\textbf{Mindvis} & 0.1921$_{\pm 0.02}$  & 0.4304$_{\pm 0.03}$  & 0.697$_{\pm 0.04}$   & 0.7162$_{\pm 0.03}$ & 0.6119 $_{\pm 0.03}$ \\

\textbf{Mindeye1} & 0.2286 $_{\pm 0.03}$  & 0.4766 $_{\pm 0.02}$ & 0.6807 $_{\pm 0.05}$ & 0.8093 $_{\pm 0.01}$ &0.8002 $_{\pm 0.04}$  \\
\textcolor{black}{\textbf{MindBridge}  }   & \textcolor{black}{0.1898$_{\pm 0.04}$} & \textcolor{black}{0.4227$_{\pm 0.02}$} & \textcolor{black}{0.6960$_{\pm 0.04}$} & \textcolor{black}{0.8219$_{\pm 0.01}$} & \textcolor{black}{0.7960$_{\pm 0.01}$}

\\
\textcolor{black}{\textbf{NeuralDiffuser}} & \textcolor{black}{0.2006$_{\pm 0.02}$} & \textcolor{black}{0.4253$_{\pm 0.03}$} & \textcolor{black}{0.6836$_{\pm 0.02}$} & \textcolor{black}{0.8501$_{\pm 0.01}$} & \textcolor{black}{0.8372$_{\pm 0.03}$}

\\
\textbf{Mindeye2} &0.2442 $_{\pm 0.03}$  & 0.4952 $_{\pm 0.01}$ & 0.6586 $_{\pm 0.02}$  & 0.8322$_{\pm 0.02}$ & 0.8280$_{\pm 0.04}$ \\
\midrule
\textbf{\modelname}+SDXL  &\textcolor{red}{0.2669 $_{\pm 0.01}$}  & \textcolor{red}{0.5537 $_{\pm 0.01}$} & \textcolor{red}{0.5989 $_{\pm 0.03}$}  & \textcolor{red}{0.8727$_{\pm 0.03}$} & \textcolor{red}{0.8820$_{\pm 0.04}$} \\
\textbf{Mde2}+SDXL &0.2500 $_{\pm 0.04}$  & 0.5511 $_{\pm 0.04}$ & 0.6224 $_{\pm 0.01}$  & 0.8678$_{\pm 0.02}$ & 0.8556$_{\pm 0.01}$ \\
\bottomrule

\end{tabular}
}
\label{tb:main}
\end{table}

\begin{figure}[t]
    \centering
    \includegraphics[width=0.9\linewidth]{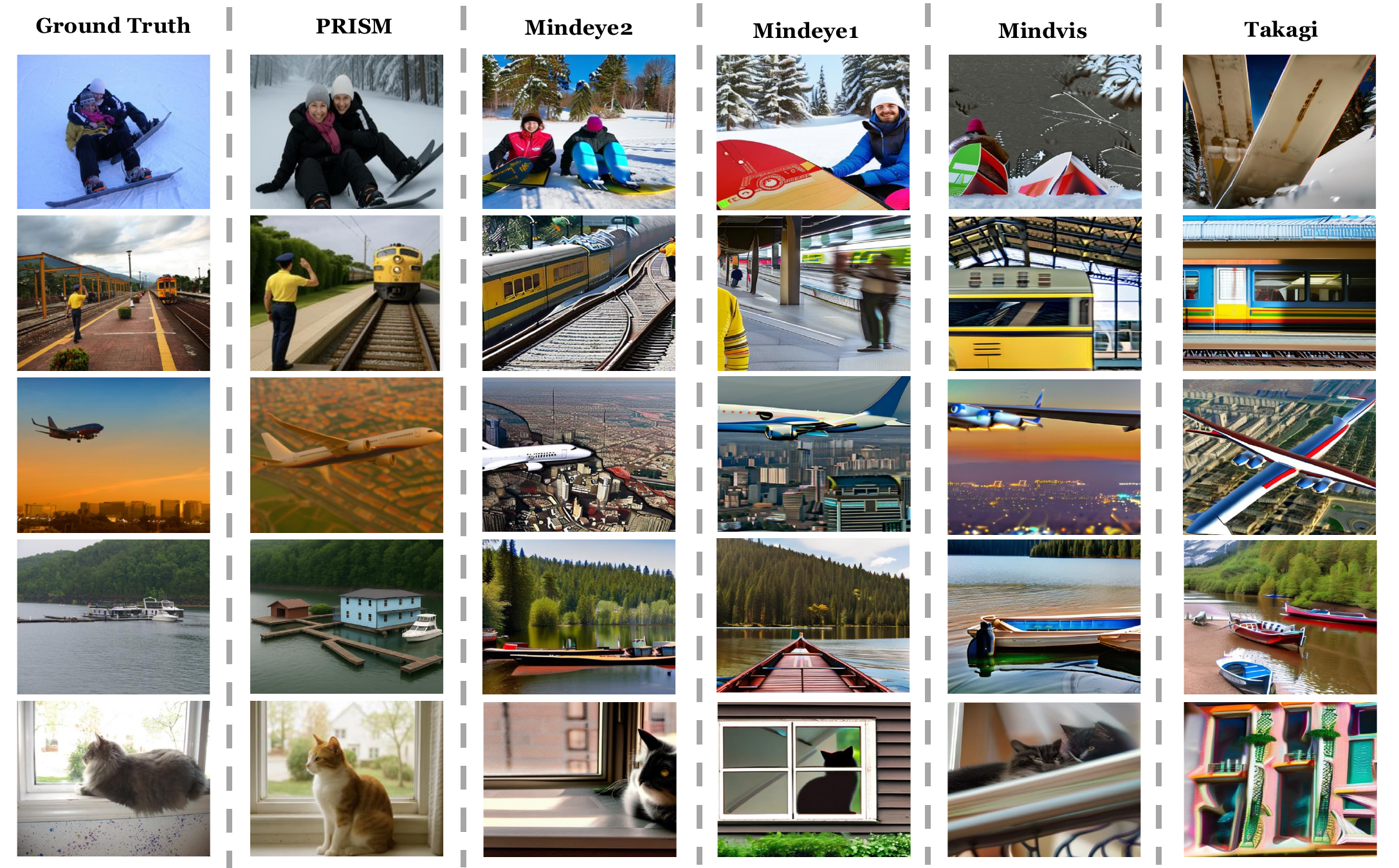}
    \vspace{-0.3cm}
    \caption{\small Reconstructed images from different methods. The first column shows the original viewed images. The rest of the columns show the reconstructed images from different methods. }
    \label{fig:reconstructed}
\end{figure}

We evaluate our fMRI-to-image reconstruction framework on test data and compare it with state-of-the-art methods.

\begin{wraptable}{r}{0.55\textwidth}    
\vspace{-0.75cm}
\centering
\caption{\small Image QA results on reconstructed NSD images. \textbf{Mdvs}, \textbf{Mde1} and \textbf{Mde2} refer to \textbf{Mindvis}, \textbf{Mindeye1} and \textbf{Mindeye2}, respectively. Results are reported as accuracy, with the best highlighted in \textcolor{red}{red}.}
    \resizebox{0.9\linewidth}{!}{

\begin{tabular}{lccccc}
\toprule

    & \textbf{\modelname}   & \textbf{Takagi} & \textbf{Mdvs} & \textbf{Mde1} & \textbf{Mde2} \\
\midrule    
\textbf{Acc.}$\uparrow$ & \textcolor{red}{0.6054} & 0.4011    & 0.5037  & 0.5516   & 0.5765  \\
\bottomrule
\end{tabular}
}
\label{tb:qa}
\vspace{-0.2cm}
\end{wraptable}

 Table \ref{tb:main} summarizes the results, with all metrics and standard deviations averaged across subjects over five runs. Visualizations of reconstructed examples from the test set are shown in Figure~\ref{fig:reconstructed}. As shown in Table \ref{tb:main}, \modelname outperforms state-of-the-art methods across all datasets and metrics, with up to a 6\% improvement in LPIPS, indicating higher perceptual similarity to the original images. Unlike baselines such as Mindeye2~\citep{scotti2024mindeye2} and Mindeye1~\citep{scotti2023reconstructing}, which often ignore key objects, \modelname successfully reconstructs all objects, yielding notable gains across all metrics. These results demonstrate the effectiveness of our framework in translating brain activity into accurate, perceptually aligned image reconstructions.
%The substantial improvements further highlight that our approach generates images that are not only structurally accurate but also perceptually closer to the original images, emphasizing the advantages of object-centric fMRI mapping and object diffusion in enhancing reconstruction quality. 

We further evaluate our method on a question answering (QA) task using reconstructed test images from the NSD dataset. For each image, we retrieve a corresponding question–answer pair from the COCO dataset~\citep{lin2014microsoft} and use Qwen2.5~\citep{bai2025qwen2} to answer the question based on the generated image. QA accuracy is reported in Table~\ref{tb:qa}. Our method achieves an accuracy of 60.54\%, significantly outperforming state-of-the-art methods. This demonstrates that our reconstructions are not only visually faithful but also semantically meaningful.

% We further conduct image question \& answering task on the reconstructed images on NSD dataset with Qwen2.5 \citep{bai2025qwen2}. The results are measured by the accuracy of getting the correct answer. The results are shown in Table \ref{tb:qa}. Our method achieves 60.54\% accuracy, outperforming the state-of-the-art methods by a large margin. This indicates that our method is capable of reconstructing images that are not only visually similar to the original images but also semantically meaningful.

\subsection{Ablation Study}

In this subsection, we present ablation studies to justify our choice of text as the latent space and to evaluate the effectiveness of the object-centric diffusion and attribute/relationship search modules.
% evaluate \modelname's performance across different embedding variants and the contribution of each component in our framework to fMRI-to-image reconstruction performance. 
Experiments are conducted on the NSD dataset, though the trend generalizes to other datasets. 

\begin{wraptable}{r}{0.6\textwidth}
    
\vspace{-0.35cm}
\centering
\caption{\small Reconstruction performance across three latent spaces. The best result in each column is highlighted in \textcolor{red}{red}. $\uparrow$ indicates higher is better and $\downarrow$ indicates lower is better.}
\vspace{0.1cm}
\resizebox{0.6\textwidth}{!}{
\begin{tabular}{lccccc}
\toprule
          
    \textbf{\small NSD}      & \textbf{PixCorr} $\uparrow$ & \textbf{SSIM} $\uparrow$  & \textbf{LPIPS} $\downarrow$  & \textbf{CLIP} $\uparrow$  & \textbf{Inception V3} $\uparrow$  \\ 

          %\multicolumn{5}{l}{\textbf{\small NSD}} &   \\
\midrule
\textbf{\modelname}      & \textcolor{red}{0.3404$_{\pm 0.05}$}  & \textcolor{red}{0.4640$_{\pm 0.02}$}  & \textcolor{red}{0.5963$_{\pm 0.02}$} &     \textcolor{red}{0.9467$_{\pm 0.03}$}   & \textcolor{red}{0.9516$_{\pm 0.03}$}  \\
\textbf{Clip text} & 0.3208$_{\pm 0.04}$    & 0.3725$_{\pm 0.06}$ & 0.6611$_{\pm 0.05}$ &0.9197$_{\pm 0.02}$  &  0.9011$_{\pm 0.04}$  \\
\textbf{LDM}   & 0.2090$_{\pm 0.07}$  & 0.3727$_{\pm 0.07}$ & 0.7502$_{\pm 0.04}$ & 0.8602 $_{\pm 0.06}$ & 0.8925 $_{\pm 0.05}$ \\
\bottomrule
\end{tabular}
\label{tb:abl_diff_space}
}
\vspace{-0.3cm}
\end{wraptable}

We first compare reconstruction performance across three latent spaces: (1) language model embeddings (ours), (2) CLIP text embeddings (CLIP-Text), and (3) the image latent space of a diffusion model (LDM). As shown in Table~\ref{tb:abl_diff_space}, aligning fMRI signals to the language model text space consistently outperforms the other two spaces across all metrics. This demonstrates that textual representations alone can capture multiple levels of visual information, making text space a more brain-aligned and effective intermediate representation for fMRI-to-image reconstruction. Results for the remaining two datasets are provided in Table~\ref{tb:app_abl_space} of the Appendix.

\begin{wraptable}{r}{0.6\textwidth}    
\vspace{-0.7cm}
    \centering
    \caption{\small Effectiveness of the object-centric diffusion module and attribute/relationship search module on NSD data. The best result is highlighted in \textcolor{red}{red}.}
\vspace{0.1cm}
\resizebox{0.8\linewidth}{!}{
\begin{tabular}{lccc}
\toprule

               & \textbf{PixCorr} $\uparrow$ & \textbf{SSIM} $\uparrow$  & \textbf{LPIPS} $\downarrow$ \\
\midrule

\textbf{\modelname}         & \textcolor{red}{0.3404$_{\pm 0.05}$}  & \textcolor{red}{0.4640$_{\pm 0.02}$}  & \textcolor{red}{0.5963$_{\pm 0.05}$} \\
\textbf{w/o ObjC.}     & 0.3291 $_{\pm 0.06}$  & 0.4299 $_{\pm 0.06}$ & 0.6111 $_{\pm 0.05}$ \\
\textbf{w/o AttOpt.+Bst}   & 0.3311$_{\pm 0.04}$    & 0.4421$_{\pm 0.01}$ & 0.6005$_{\pm 0.02}$ \\
%-t5+llama 2 7b & 0.3011  & 0.4094 & 0.6606\\
% -prompt opt + random select &     &  & \\ 
\textbf{w/o AttOpt.+Wst} &  0.3068 $_{\pm 0.05}$    & 0.4167$_{\pm 0.02}$  & 0.6398$_{\pm 0.05}$  \\ 

\bottomrule
\end{tabular}}
\label{tab:ablation}
\vspace{-0.2cm}
\end{wraptable}

Next, we evaluate the effectiveness of the two proposed modules. Results are in Table~\ref{tab:ablation}. To evaluate the Object-centric Diffusion module, we compare against a variant (\textbf{w/o ObjC.}) that replaces object-level cross-attention with standard U-Net cross-attention. To assess the attribute/relationship search module, we test two variants that skip the search process and rely only on the initial keyword set: \textbf{w/o AttOpt.+Bst}, which fixes the prompt to the highest-scoring (best) keyword, and \textbf{w/o AttOpt.+Wst}, which fixes it to the lowest-scoring (worst) keyword.
%The initial keywords are provided in Appendix~\ref{app:attr_opt}.

Overall, removing or replacing the two modules consistently degrades performance across all metrics. Specifically, eliminating object cross-attention leads to notable declines that cannot be recovered through prompt optimization, highlighting its essential role in reconstructing perceptually accurate images. Likewise, bypassing prompt optimization and using the best or worst initial attribute also reduces performance, indicating that the initial attributes alone are insufficient and underscoring the importance of prompt optimization in our model. The ablation study on the number of objects in our framework is shown in~\Cref{app_mlp} of the Appendix.

\subsection{Keyword Search}
\textbf{Case Study.} To better understand the keywords selected by our attribute/relationship search module, Table~\ref{tab:search_words} presents the top-scoring keywords across different rounds of the $\varepsilon$-greedy search.
% In this subsection, we describe the experimental setup for our attribute/relationship search, which aims to identify the key attributes and relationships encoded by brain signals during visual processing.
% The optimization is conducted on the training data using an $\varepsilon$-greedy search strategy with $\varepsilon = 0.5$, allowing a balance between exploration and exploitation over 40 update rounds. At each search round, we prompt the VLM to generate two new relationship keywords based on eight words in the current keyword set. 
% The search is initialized with six widely-used keywords describing object relationships: Semantic Relation~\citep{johnson2015image}, Positional Relation~\citep{lu2016visual,haldekar2017identifying}, Functional Relation~\citep{zhu2015understanding}, Action Relation~\citep{lu2016visual}, Visual Attributive Relation~\citep{farhadi2009describing}, and Part–Whole Relation\citep{lu2016visual}. For each type, we use \texttt{GPT-4o} to generate four synonymous keywords, resulting in an initial pool of 24 candidate attributes. We adopt the scoring function $\mathcal{S}(\mathbf{Y}_1,\mathbf{Y}_2) = 1 - \text{LPIPS}(\mathbf{Y}_1,\mathbf{Y}_2)$ as defined in Equation~\ref{eq:opt_obj}.  The CKA threshold $\alpha$ is initialized to the minimum CKA score among the initial candidates. 
The results show that, despite extensive exploration, the top-scoring keywords consistently converge toward spatially oriented relationships such as \texttt{Spatial Layout} and \texttt{Relative Position}. This suggests that: (1) descriptions emphasizing spatial information are most effective for guiding the diffusion model to accurately reconstruct images, as indicated by their highest LPIPS scores; and (2) these keywords also align well with fMRI data, as their CKA scores are no lower than those of the initial keywords, in accordance with the search constraints. This result is consistent with prior neuro-scientific findings showing that neural representations in the brain are sensitive to spatial arrangements and relative positions of objects~\citep{zopf2018representing, graumann2022spatiotemporal}. Therefore,  we use \texttt{Spatial Layout} as the optimal keyword $a$ to generate structured descriptions for model training.
{Additional implementation details of the attribute/relationship search module can be found in Section~\ref{app:attr_opt} of the Appendix. To further demonstrate its robustness, we perform an experiment where all spatial-related keywords are deliberately excluded from the initial keyword set. The results, presented in Section~\ref{app:new_keyword_search} of the Appendix, show that the module is still able to identify the relevant spatial keywords, demonstrating the robustness and effectiveness of our search process.
}

{
\color{black}
\textbf{Neuroscientific Interpretation of the Optimal Keyword.} To explore why the optimal keyword \texttt{Spatial Layout} exhibits stronger alignment with fMRI signals, we conduct a gradient-based interpretability analysis to examine the neural correlations of different keyword types. Specifically, during inference on the test set, we identify predicted words associated with spatial relationships (e.g., directional terms) and functional attributes (e.g., actions or behaviors of objects), and compute the gradients of these tokens with respect to the input fMRI signals. Ideally, larger gradients indicate that specific voxel values within the fMRI contribute more strongly to predicting the given keyword, and are thus more relevant to that semantic category. By averaging these gradients across samples, we identify the ROIs that contribute most to each keyword category. For spatial relationships, we observe the strongest activation in the ROI 119 (HCP mmp1 atlas)~\citep{glasser2016multi}, which belongs to Presubiculum (PreS), a region linked to spatial memory~\citep{dalton2017pre, boecker2024hippocampal}. In contrast, for functional attributes, the Ventromedial Visual Area 1 (VMV1, ROI 153 of HCP mmp1 atlas) shows the highest activation. Additionally, we compute the mean voxel intensity within each ROI and find that PreS exhibited a higher mean activation (0.0080) compared to VMV1 (0.0028). These findings suggest that the fMRI contains stronger activation in spatially relevant regions such as PreS, which may explain the higher alignment observed for spatial keywords. Nonetheless, we acknowledge that further validation from the neuroscience community is needed.
}

%% file: 05relatedworks.tex
\section{Related Work}

\textbf{FMRI-Image Reconstruction.} Early approaches leveraged linear models to decode fMRI signals into visual features~\citep{kay2008identifying, takagi2023high}. Later work explores deep generative models: approaches such as~\citep{lin2022mind, ozcelik2022reconstruction, goodfellow2020generative} map fMRI signals into the latent space of GANs for image reconstruction, while \textcolor{black}{BraVL~\citep{du2023decoding} employs a multimodal VAE to jointly model relationships between brain activity and visual–linguistic features for neural decoding}. With advances in vision–language models~\citep{radford2021learning, liang2024survey}, more recent approaches map fMRI signals into CLIP’s image-embedding space~\citep{scotti2024mindeye2, scotti2023reconstructing} and then use diffusion models for image reconstruction~\citep{rombach2022high, xu2023versatile, podell2023sdxl}. \textcolor{black}{Building on this diffusion-based direction, MindDiffuser~\citep{lu2023minddiffuser} introduces a two-stage framework that first decodes semantic information and then aligns structural information with CLIP visual features decoded from fMRI.} Unlike prior work that directly maps fMRI signals to joint text–image spaces~\citep{wang2024decoding, quan2024psychometry}, we compare multiple representation spaces and find that text embeddings from language models~\citep{raffel2020exploring} exhibit the strongest alignment with fMRI signals. This insight motivates our approach of reconstructing images via the embedding space of language models. \textcolor{black}{Cross-subject decoding has also gained attention. MindTuner~\citep{gong2025mindtuner} introduces visual-fingerprint modeling using Skip-LoRA to improve cross-subject alignment and semantic correction.
Psychometry~\citep{quan2024psychometry} proposes an omni Mixture-of-Experts architecture that captures both inter-subject shared structure and subject-specific variability, combined with retrieval-enhanced inference for improved reconstruction.
Wills Aligner~\citep{bao2025wills} further advances multi-subject collaborative decoding by combining anatomical alignment with subject-guided Mixture-of-Brain-Expert adapters. Beyond static images, recent work has extended brain decoding to dynamic visual stimuli.
NeuroClips~\citep{gong2024neuroclips} reconstructs such stimuli from fMRI by combining decoded keyframes with low-level perceptual flows to improve temporal consistency.
BrainNEDS~\citep{yeung2025reanimating} utilizes video diffusion models to disentangle static and dynamic components, offering insights into which visual features are prioritized by the brain.}

% \textbf{Diffusion models} have been widely adopted across a range of tasks, including image generation/editing~\citep{wijmans1995solution, gal2022image, song2020denoising}, and text-to-image synthesis~\citep{ruiz2023dreambooth}.  These generative models learn to synthesize data by reversing a predefined noise-injection process. To enhance image quality, several techniques have been proposed. For example, ControlNet \citep{zhang2023adding} specify some high-level features of images for controlling semantic structures and, and GLIGEN \citep{li2023gligen} design position-aware adapters on top of the diffusion models for spatially-conditioned image generation. Another approach is to use training-free methods to steer diffusion models through manipulating latent or cross-attention maps according to spatial or semantic constraints during inferences without additional training \citep{chen2024training, yang2024mastering}.

\textbf{Diffusion models.} Diffusion models have become foundational in generative tasks like image creation and editing~\citep{gal2022image, song2020denoising}, as well as text-to-image synthesis~\citep{ruiz2023dreambooth}. To enhance control over generated content, ControlNet~\citep{zhang2023adding} introduces high-level image features for controlling and GLIGEN~\citep{li2023gligen,zhang2025layercraft} incorporates position-aware adapters for spatial grounded generation. Meanwhile, there are also training-free methods that adjust latent or attention maps during inference to guide outputs without additional training~\citep{chen2024training, yang2024mastering}. In our work, we guide the diffusion process by modifying cross-attention layers during inference to integrate object-level descriptions derived from fMRI data for image reconstruction.

\textbf{Prompt Optimization.} Prompt optimization aims to discover effective textual prompts for LLMs without model fine-tuning. Gradient-based methods~\citep{shin2020autoprompt, shi2022toward, wen2023hard} update prompts using gradients or differentiable embeddings. Gradient-free approaches treat LLMs as black boxes, using heuristic search~\citep{prasad2022grips, pryzant2023automatic}, reinforcement learning~\citep{deng2022rlprompt, zhang2022tempera}, or evolutionary strategies~\citep{zhou2022large, yang2023large, guo2025evopromptconnectingllmsevolutionary}. We designed our gradient-free prompt optimization based on beam search to optimize attribute keywords for black-box vision-language models.

\begin{table}[t]
\centering
\scriptsize
\caption{\small Top-5 keywords scored by $1-\text{LPIPS}$ before searching and after 10, 20, 30 search steps. The top-5 results remain unchanged after 30 search rounds. The search results indicate a clear preference for keywords related to spatial and positional relations, with most of the top-performing keywords in the final results containing the term '\textit{spatial}'.}
\resizebox{\linewidth}{!}{
\begin{tabular}{c|cccc}
\toprule
\normalsize{\textbf{Rank}} & \normalsize{\textbf{Initial}} & \normalsize{\textbf{Round 10}} & \normalsize{\textbf{Round 20}} & \normalsize{\textbf{Round 30}+}\\ \hline
\normalsize{\#1} & \texttt{Spatial Configuration} & \texttt{Spatial Arrangement} & \texttt{Spatial Organization} & \texttt{Spatial Layout} \\
\normalsize{\#2} & \texttt{Positional Relation} & \texttt{Spatial Configuration} & \texttt{Spatial Structure} & \texttt{Spatial Patterns} \\
\normalsize{\#3} & \texttt{Location Relation} & \texttt{Spatial Interaction} & \texttt{Spatial Arrangement} & \texttt{Spatial Organization} \\
\normalsize{\#4} & \texttt{Descriptive Attribute} & \texttt{Positional Relation} & \texttt{Spatial Configuration} & \texttt{Relative Position} \\
\normalsize{\#5} & \texttt{Inclusion Dependency} & \texttt{Feature Relation} & \texttt{Spatial Interaction} & \texttt{Spatial Relationships} \\
\bottomrule
\end{tabular}
}
\label{tab:search_words}
\end{table}

%% file: 06conclusion.tex
\section{Conclusion}

% In this work, we addressed the challenge of reconstructing visual stimuli from fMRI signals. Our analysis shows that fMRI signals align more closely with language model text space than with visual-based alternatives, motivating text as a brain-aligned intermediate representation. Moreover, we find that using the compositional nature of visual perception further improves reconstruction quality when integrated with generation models. Motivated by these finds, we introduce \modelname, a framework that projects fMRI signals into a structured text space and incorporates an object-centric diffusion module and an attribute/relationship search module. Extensive experiments on real-world fMRI datasets demonstrate that PRISM outperforms prior methods, achieving up to an $8\%$ reduction in perceptual loss.
In this work, we address the challenge of reconstructing visual stimuli from fMRI signals. Our analysis reveals that fMRI signals align more closely with the text space of language models than with vision-based or joint text–image representations, identifying text as a brain-aligned intermediate space. Building on this insight, we show that explicitly modeling the compositional structure of visual perception—capturing objects along with their attributes and relationships—further improves reconstruction quality. Guided by these findings, we develop \modelname, a framework that maps fMRI signals into a structured text space and incorporates two specialized modules: an object-centric diffusion module that generates images by composing individual objects, and an attribute/relationship search module that automatically discovers attributes and relationships aligned with neural activity. Experiments on real-world fMRI datasets demonstrate that PRISM reduces perceptual loss by up to 6\% compared to prior methods, underscoring the power of structured text as a bridge between brain activity and image generation.

\section{Ethics statement}
Our work does not involve human or animal subjects, personally identifiable data, or sensitive information. The datasets used are publicly available, and we follow their respective licenses. The methods and findings presented do not pose foreseeable risks of misuse, discrimination, or harm. We therefore believe our work raises no specific ethical concerns under the ICLR Code of Ethics.

\section{Reproducibility statement}

\Cref{sec:method} details the proposed framework and its design. \Cref{sec:exp} describes the datasets, baseline methods, and evaluation protocols used for comparison. Additional implementation details, including training procedures and hyperparameter settings, are provided in the Appendix. Upon acceptance of this paper, we will release our code on GitHub.

%% file: 10appendix.tex
\newpage

\section{A Common Error in Generative Models}
\label{app:common_errors}
\begin{figure}[ht]
    \centering
    \includegraphics[width=0.8\linewidth]{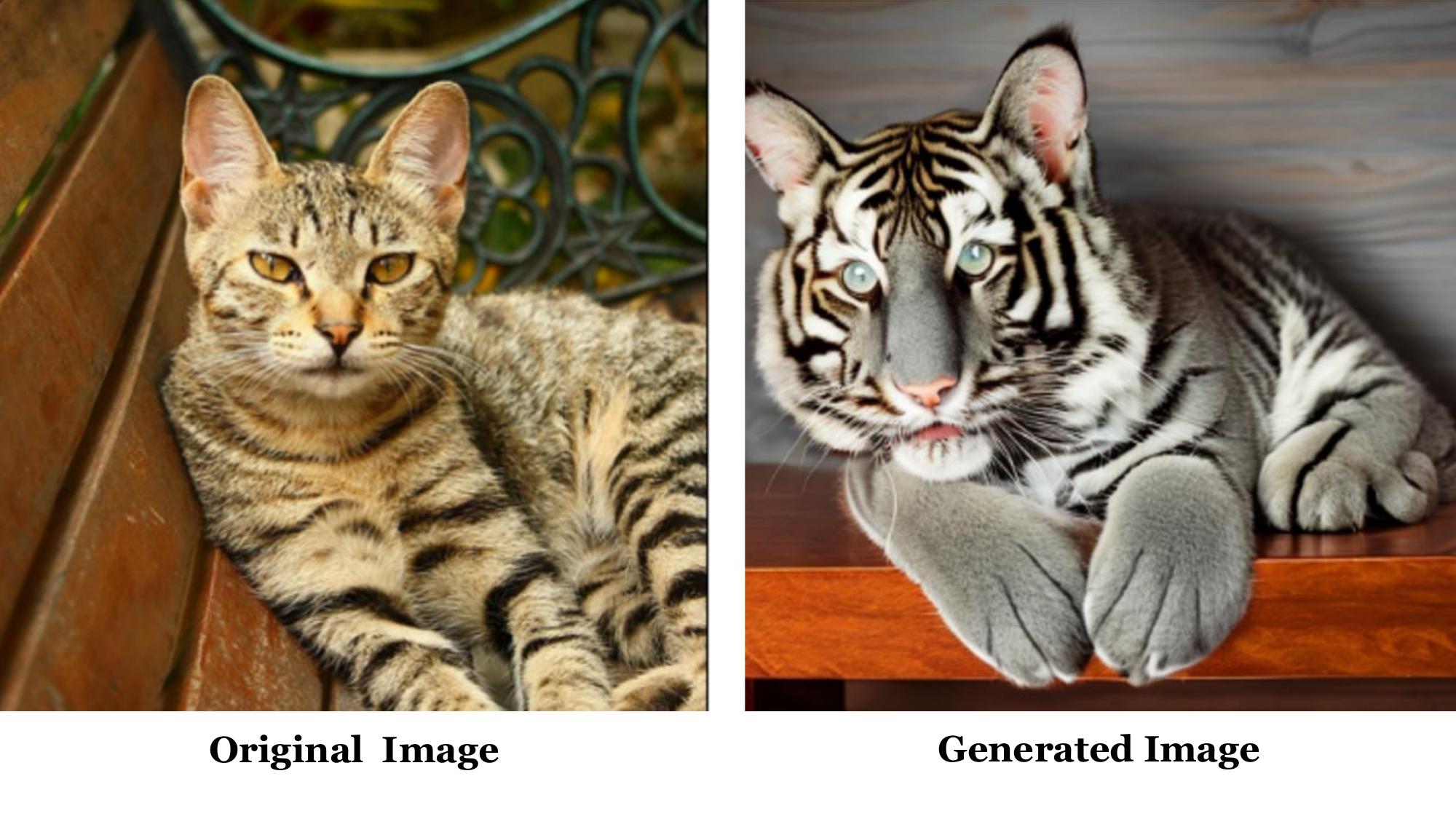}
    \caption{A Common Error in Generative Models. While the original image shows “a gray tiger-striped cat,” the model incorrectly generates “a grey tiger,” illustrating semantic distortion. }
    \label{fig:common_errors}
\end{figure}

In this section, we present a common failure, attribute binding, encountered in generative models (especially for the diffusion model), where generative models misattribute visual properties to objects. Figure~\ref{fig:common_errors} compares original (left) and generated (right) images. The original depicts a gray tiger-striped cat on a wooden bench, while the generated version incorrectly shows a gray tiger instead of a cat. This issue arises because diffusion models usually rely on text encoders such as CLIP, which are known to lack the ability to capture complex linguistic structures~\citep{yuksekgonul2022and}. Consequently, the diffusion process loses awareness of the bindings between objects and their attributes, leading to mismatched visual properties. This impairs fMRI-to-image reconstruction. To address this, we introduce a neuroscience-inspired, object-centric generation approach that improves reconstruction quality.

\section{Preliminary: Diffusion Model}
\label{app:diffusion}

Diffusion models are a class of generative models that synthesize data by reversing a gradual noising process. Given a data point $ \mathbf{H}_0$ (e.g., an image), the forward process perturbs it into Gaussian noise over $T$ time steps. The model then learns the reverse process to reconstruct samples from noise. The forward process is a Markov chain defined by:
\begin{equation*}
q(\mathbf{H}_t \mid \mathbf{H}_{t-1}) = \mathcal{N}(\mathbf{H}_t; \sqrt{1 - \beta_t} \, \mathbf{H}_{t-1}, \beta_t \mathbf{I}), \quad t = 1, \ldots, T,
\end{equation*}

where \( \{ \beta_t \}_{t=1}^T \) is a predefined variance schedule. The model is trained to predict the noise \( \boldsymbol{\epsilon} \) added to the input, using a neural network \( \boldsymbol{\epsilon}_\theta \), by minimizing:

\begin{equation*}
\mathcal{L} = \mathbb{E}_{\mathbf{H}_0, \boldsymbol{\epsilon}, t} \left[ \left\| \boldsymbol{\epsilon} - \boldsymbol{\epsilon}_\theta(\mathbf{H}_t, t) \right\|^2 \right]. 
\end{equation*} 

Here, \( \mathbf{H}_t = \sqrt{\bar{\alpha}_t} \, \mathbf{H}_0 + \sqrt{1 - \bar{\alpha}_t} \, \boldsymbol{\epsilon} \), with \( \bar{\alpha}_t = \prod_{s=1}^t (1 - \beta_s) \), and \( \boldsymbol{\epsilon} \sim \mathcal{N}(0, \mathbf{I}) \). To guide the generation process with external information \( \mathbf{C} \) (e.g., a text prompt), the denoising network is extended as:
\begin{equation*}
\boldsymbol{\epsilon}_\theta(\mathbf{H}_t, t, \mathbf{C}).
\end{equation*} Then, the training objective becomes:

\begin{equation*}
\mathcal{L}_{\text{cond}} = \mathbb{E}_{\mathbf{H}_0, \boldsymbol{\epsilon}, t, \mathbf{C}} \left[ \left\| \boldsymbol{\epsilon} - \boldsymbol{\epsilon}_\theta(\mathbf{H}_t, t, \mathbf{C}) \right\|^2 \right].
\end{equation*}
This formulation is widely used in text-to-image diffusion models, where \( \mathbf{C} \) is the embedding of a textual description obtained from a pre-trained text encoder (e.g., CLIP). In practice, the condition \( \mathbf{C} \) is incorporated into the U-Net via cross-attention modules. 
%Each attention block computes:

% \begin{equation*}
% \text{CrossAttention}(\mathbf{Q}, \mathbf{K}, \mathbf{V}) = \text{softmax} \left( \frac{\mathbf{Q} \mathbf{K}^\top}{\sqrt{d}} \right) \mathbf{V},
% \end{equation*}

% where \( \mathbf{Q} = \mathbf{W}_Q \cdot \phi(\mathbf{h}_t) \) are queries from the image features, and \( \mathbf{K} = \mathbf{W}_K \cdot\varphi(\mathbf{c}) \), \( \mathbf{V} = \mathbf{W}_V \cdot \varphi(\mathbf{c}) \) are keys and values derived from the condition \( \mathbf{c} \). Here, \( \phi(\cdot) \) and \( \varphi(\cdot) \) are feature projections (e.g., linear layers, layer normalization or non-linearities depending on implementation), and $ d $ is the feature dimension. This mechanism allows the model to attend to the conditioning input \( \mathbf{c} \) at each step of generation, aligning visual outputs with semantic content.

In our work, we adopt the latent diffusion framework~\citep{rombach2022high}, where the diffusion process is applied in the latent space of a pre-trained VAE, rather than directly in pixel space. Specifically, an input image \( \mathbf{Y} \in \mathbb{R}^{H \times W \times 3} \) is first encoded by a VAE encoder into a compact latent representation \( \mathbf{Z} \in \mathbb{R}^{h \times w \times c} \):
\[
\mathbf{Z} = \text{Encoder}(\mathbf{Y}).
\]
The diffusion process is applied on \(\mathbf{Z}\), where the perturbed latent representation \(\mathbf{Z}_T\) is obtained after T steps. The reversed denoising steps then generate a denoised latent \(\hat{\mathbf{Z}}\) over T steps. The final image is reconstructed by the denoised latent:
\[
\hat{\mathbf{Y}} = \text{Decoder}(\hat{\mathbf{Z}}).
\]

This formulation greatly reduces computational cost while maintaining high-quality image generation and is particularly well-suited for conditioning on high-level semantic representations such as text or fMRI-derived embeddings.

\section{Dataset}
\label{app:dataset}

In this subsection, we provide information about the three pre-processed datasets used for the fMRI-to-image reconstruction task: NSD \citep{allen2022massive}, BOLD5000 \citep{chang2019bold5000}, and GOD \citep{horikawa2017generic}. 

\begin{itemize}
    \item \textbf{NSD} \citep{allen2022massive}: The Natural Scenes Dataset (NSD) is a large-scale public fMRI dataset capturing brain responses of human participants viewing naturalistic stimuli from COCO images \citep{lin2014microsoft}. The dataset includes scans for 30–40 hours across 30–40 separate sessions. During each session, participants viewed 750 images for 3 seconds each. Each image was presented three times across sessions, with most images unique to each subject, except for 1,000 shared images seen by all subjects. Following prior NSD reconstruction studies \citep{scotti2023reconstructing, takagi2023high}, we adopt the standardized train/test split, where the shared images serve the test set. Consequently, the training set for each subject contains 8,859 image stimuli and 24,980 fMRI trials, while the test set includes 982 image stimuli and 2,770 fMRI trials.
    
    \item \textbf{BOLD5000} \citep{chang2019bold5000}: The BOLD5000 dataset is a publicly available fMRI dataset capturing brain activity as subjects viewed a series of images. It contains 4,916 unique images, including 2,000 from the COCO dataset and 1,916 from ImageNet \citep{deng2009imagenet}. Each image was presented as a visual stimulus in individual trials. Of these, 4,803 images were shown once, while 113 images were repeated three or four times across trials, resulting in a total of 5,254 stimulus trials. We follow the standardized train/test split used in prior BOLD5000 reconstruction studies \citep{chen2023seeing, wang2024decoding}. Specifically, the training set includes trials with non-repeated image stimuli, comprising 4,803 samples, while the test set consolidate repeated image stimulus trials into 113 samples.
    
    \item \textbf{GOD} \citep{horikawa2017generic}: The Generic Object Decoding (GOD) is a public dataset developed for fMRI based decoding. It aggregates fMRI data gathered through the presentation of images from 200 representative object categories, originating from ImageNet. We follow the standardized train/test set split employed in existing GOD image reconstruction studies \citep{sun2023contrast} and get 1200 training samples and 50 test samples.The Generic Object Decoding (GOD) dataset is a publicly available fMRI dataset designed for decoding object representations. It includes fMRI data collected during the presentation of images from 200 representative object categories sourced from ImageNet. Following the standardized train/test split used in prior GOD reconstruction studies \citep{sun2023contrast}, we use 1,200 training samples and 50 test samples.
\end{itemize}

\section{Baselines}
\label{app:baseline}

In this section, we provide details about the baselines used in our experiments.

\begin{itemize}
    \item Takagi \& Nishimoto \citep{takagi2023high}: This baseline maps fMRI signals to the latent space of a pre-trained VAE within a diffusion model using linear regression, enabling image reconstruction. The method combines image latent representations with text embeddings extracted from a CLIP text encoder, both mapped from fMRI signals in higher (ventral) visual cortex regions, to improve reconstruction quality. For a fair comparison, we adapt this approach to the Diffusion 2.1 pipeline by retraining the linear regression to map fMRI signals to both the VAE latent space and the textual conditioning used in Diffusion 2.1.
            
    \item Mindvis \citep{chen2023seeing}: This baseline uses a self-supervised representation of fMRI data using masked modeling within a high-dimensional latent space in an encoder–decoder framework. The learned representation is then projected into the conditioning space of LDM by fine-tuning the model. For fair comparison, we adapt the image generator of this approach to Diffusion 2.1 by fine-tuning it with the learned projection module, following the strategy outlined by the original authors.

    \item Mindeye \citep{scotti2023reconstructing}: This model proposed two modules to map the fMRI signal to the CLIP image space. Specifically, the model first uses contrastive learning to align fMRI signals with image embeddings. Second, the paper trains a diffusion prior to reconstructing images from these embeddings via mapping brain activity into CLIP image space, enabling the generation of images that closely resemble the original stimuli. To adapt the method for fair comparison, we replace the Versatile Diffusion with Diffusion 2.1.
    {\color{black}
    \item MindBridge~\citep{wang2024mindbridge}: This approach unifies heterogeneous voxel dimensions using adaptive max pooling and employs a cyclic fMRI reconstruction mechanism to align neural responses across subjects in a common semantic space. To adapt the method for fair comparison, we replace the Versatile Diffusion with Diffusion 2.1.

    \item NeuralDiffuser~\citep{li2025neuraldiffuser}: This paper proposes a visual feature–guided reconstruction method that decodes multiple layers of CLIP’s visual encoder from fMRI and uses these features as gradient-based cues during the early stages of reverse diffusion, enabling bottom-up detail refinement. It further introduces momentum alignment to reduce the distribution shift between training and testing embeddings. For fair comparison, we use Diffusion 2.1 as the generative backbone.
    }
    \item Mindeye2 \citep{scotti2024mindeye2}: This method trains multiple MLPs to project fMRI signals from all subjects into a shared representation space, followed by training a diffusion prior to map these representations into the CLIP image embedding space. The final image is then reconstructed using a pre-trained SDXL~\citep{podell2023sdxl}. To adapt this method to our setting, we replace the generative backbone with Diffusion 2.1.
\end{itemize}

\section{Training Details}
\label{app:training}
\begin{figure}[H]
    \centering
    \includegraphics[width=\linewidth]{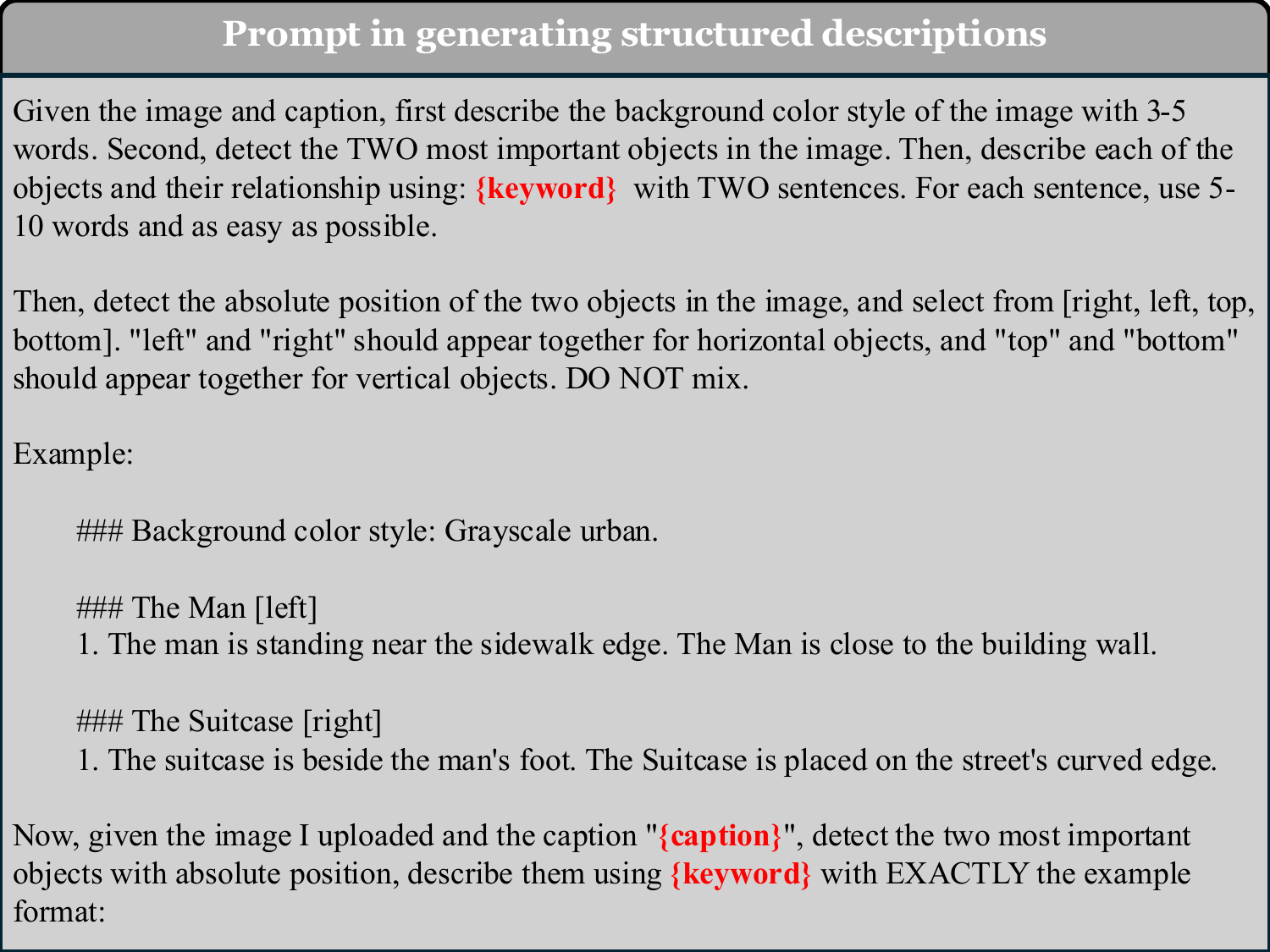}
    \label{app:fig:prompt}
    \centering
    \includegraphics[width=\linewidth]{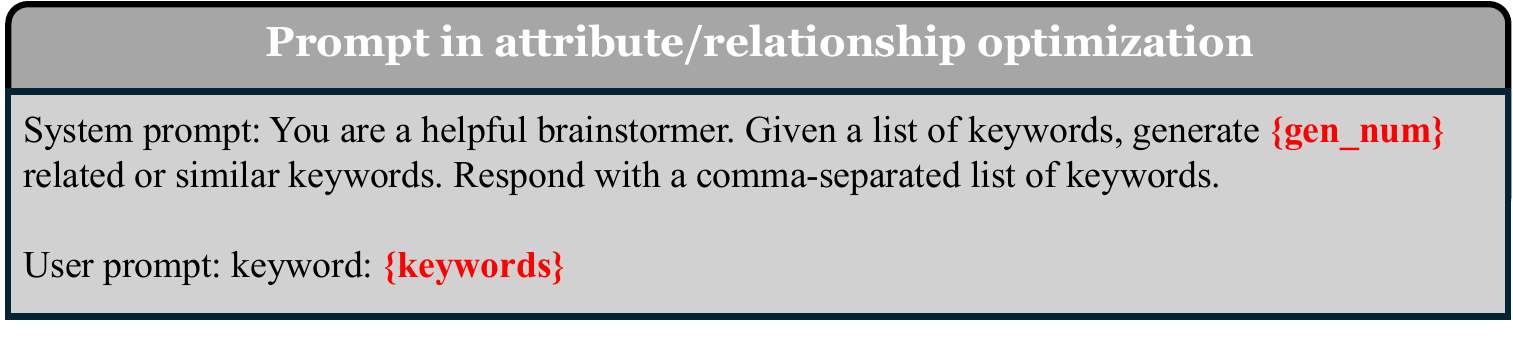}
    \label{app:fig:prompt_opt}
\end{figure}

In this section, we provide the training details of our model. Our model is implemented with Pytorch and trained on two NVIDIA-L40 GPUs with 48GB of memory. We use T5 as the language model to generate the object-level descriptions.  For NSD data, we train the model for 80 epochs: 60 epochs for MLP training ($E_{\text{MLP}}$) with a learning rate $\text{lr}_1 = 1 \times 10^{-5}$, followed by 20 epochs of joint training, where we continue training the MLP and fine-tune the T5 model ($E_{\text{T5}}$) using a learning rate $\text{lr}_2 = 5 \times 10^{-7}$. For BOLD5000, we set $E_{\text{mlp}}=50$,  $\text{lr}_1=1e-5$, $E_{\text{T5}}=5$, and $\text{lr}_2=1e^{10-8}$. For GOD, we set $E_{\text{mlp}}=40$, $\text{lr}_1=1e-5$, $E_{T5}=5$, and $\text{lr}_2=5e^{-9}$. For image reconstruction at inference time, we set the blending ratio $\beta=0.5$ and the denoising step as $40$ for all the datasets. We use \texttt{GPT 4o-mini} to generate the object-centric descriptions with the prompt shown in Section~\ref{app:fig:prompt}. For images that do not have a caption, we first use \texttt{GPT-4o-mini} to generate a short caption and then use our prompt to generate the object-centric description.

%\section{Additional Experiments}
{
\color{black}
\section{Case study on Object-Level Descriptions for Image Reconstruction}
\label{app:case_obj_level}

\begin{figure}[H]
    \centering
    \includegraphics[width=\linewidth]{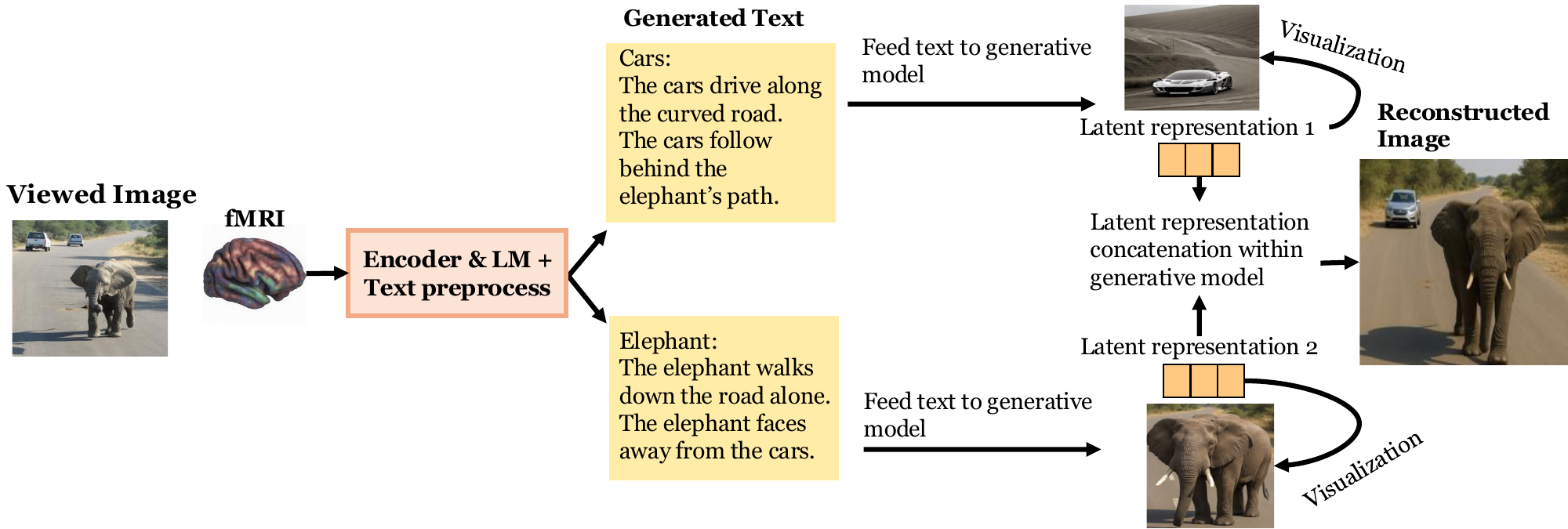}
    \caption{Reconstructed images using object-level descriptions.}
    \label{fig:obj_level}
\end{figure}

In this section, we present a case study illustrating how object-level descriptions are used during the reconstruction process, as shown in Figure~\ref{fig:obj_level}. In the given example, the generated text is segmented based on object instances: one segment describes the cars (“The cars drive along the curved road…”) and the other describes the elephant (“The elephant walks down the road alone…”). Each object description is independently fed into a diffusion-based generative model to produce object-centric latent representations. These representations are then spatially aligned and concatenated within the generative model to produce the final reconstructed image. This modular process helps mitigate common generation errors, such as attribute binding.

}

\section{Reconstruction performance across different latent spaces} 

We report the ablation studies to justify our choice of text as the latent space on BOLD5000 and GOD. As shown in Table~\ref{tb:app_abl_space}, aligning fMRI signals to the language model text space (our method) consistently outperforms alignment to the other two spaces across all metrics. This supports our core contribution: textual representations alone are sufficient to capture both high-level semantic and low-level visual information, and text space provides a more brain-aligned and effective intermediate representation for fMRI-to-image reconstruction.

\begin{table}[t] 
\vspace{-0.5cm}
\centering
\caption{Reconstruction performance across three latent spaces. The best result in each column is highlighted in \textcolor{red}{red}. $\uparrow$ indicates higher is better and $\downarrow$ indicates lower is better.}
\resizebox{\textwidth}{!}{
\begin{tabular}{lccccc}
\toprule
          
          & \textbf{PixCorr} $\uparrow$ & \textbf{SSIM} $\uparrow$  & \textbf{LPIPS} $\downarrow$  & \textbf{CLIP} $\uparrow$  & \textbf{Inception V3} $\uparrow$  \\ 

%           \multicolumn{5}{l}{\textbf{NSD}} &   \\
% \midrule
% \textbf{Ours}      & \textcolor{red}{0.3404$_{\pm 0.05}$}  & \textcolor{red}{0.4640$_{\pm 0.02}$}  & \textcolor{red}{0.5963$_{\pm 0.02}$} &     \textcolor{red}{0.9467$_{\pm 0.03}$}   & \textcolor{red}{0.9516$_{\pm 0.03}$}  \\
% \textbf{Clip text} & 0.3208$_{\pm 0.04}$    & 0.3725$_{\pm 0.06}$ & 0.6611$_{\pm 0.05}$ &0.9197$_{\pm 0.02}$  &  0.9011$_{\pm 0.04}$  \\
% \textbf{LDM}   & 0.2090$_{\pm 0.07}$  & 0.3727$_{\pm 0.07}$ & 0.7502$_{\pm 0.04}$ & 0.8602 $_{\pm 0.06}$ & 0.8925 $_{\pm 0.05}$ 
% \\ 
          \midrule

\multicolumn{5}{l}{\textbf{BOLD5000}} &   \\

\midrule

\textbf{Ours} &\textcolor{red}{0.2315$_{\pm 0.01}$}   & \textcolor{red}{0.5341$_{\pm 0.02}$}   & \textcolor{red}{0.6198$_{\pm 0.02}$}  & \textcolor{red}{0.7720$_{\pm 0.03}$} & \textcolor{red}{0.6601$_{\pm 0.07}$} \\

\textbf{Clip text} & 0.2000$_{\pm 0.06}$   & 0.4885$_{\pm 0.06}$ & 0.6520$_{\pm 0.04}$ & 0.7265$_{\pm 0.04}$ & 0.6199$_{\pm 0.04}$ \\

\textbf{LDM} & 0.1622$_{\pm 0.03}$  &0.4300$_{\pm 0.02}$  & 0.7894$_{\pm 0.08}$   & 0.7025$_{\pm 0.08}$ & 0.5590 $_{\pm 0.05}$\\

\midrule
\multicolumn{5}{l}{\textbf{GOD}} &   \\
\midrule

\textbf{Ours} & \textcolor{red}{0.2571 $_{\pm 0.01}$}   & \textcolor{red}{0.5200 $_{\pm 0.02}$} & \textcolor{red}{0.6213 $_{\pm 0.01}$} & \textcolor{red}{0.8567 $_{\pm 0.05}$} & \textcolor{red}{0.8428 $_{\pm 0.06}$} \\

\textbf{Clip text} & 0.2200 $_{\pm 0.05}$  & 0.4682 $_{\pm 0.04}$& 0.6827 $_{\pm 0.04}$ & 0.8120 $_{\pm 0.05}$ & 0.7602 $_{\pm 0.04}$ \\

\textbf{LDM} & 0.1900$_{\pm 0.02}$  & 0.3999$_{\pm 0.07}$  & 0.7100$_{\pm 0.02}$   & 0.7099$_{\pm 0.01}$ & 0.7484 $_{\pm 0.03}$ \\
\bottomrule

\end{tabular}
\label{tb:app_abl_space}

}
\vspace{0.1cm}
\end{table}

\section{Analysis on the number of objects in our framework}
\label{app_mlp}
In our framework, we fix the number of objects per image to two and assign a separate MLP to each. We learn to assign each object a location label from a predefined set of spatial positions (e.g., left/right or top/bottom). This fixed assignment of each object to a dedicated MLP, along with the predefined spatial labeling scheme, is used consistently during both training and inference. At inference time, each MLP independently encodes fMRI signals for one object, and the language model generates a structured description for each. These descriptions are then passed to the object-centric diffusion model, which generates object images independently and places them into their corresponding spatial positions to form the final image.

\begin{table}[]
\caption{Comparison of the number of objects in our framework. Results are reported using PixCorr, SSIM, LPIPS, CLIP, and Inception V3 metrics. The best result in each column is highlighted in \textcolor{red}{red}. $\uparrow$ indicates higher is better and $\downarrow$ indicates lower is better.}
\resizebox{\textwidth}{!}{

\begin{tabular}{lccccc}
\toprule
          & PixCorr $\uparrow$      & SSIM $\uparrow$         & LPIPS $\downarrow$      & CLIP $\uparrow$         & Inception V3 $\uparrow$ \\
\midrule
Ours (Two Objs) & \textcolor{red}{0.3404 $_{\pm 0.05}$} & \textcolor{red}{0.464 $_{\pm 0.02}$} & \textcolor{red}{0.5943 $_{\pm 0.02}$ }& \textcolor{red}{0.9467 $_{\pm 0.03}$} & \textcolor{red}{0.9516 $_{\pm 0.03}$} \\
One Obj   & 0.3355 $_{\pm 0.05}$ & 0.4532 $_{\pm 0.04}$ & 0.6014 $_{\pm 0.04}$ & 0.9344 $_{\pm 0.02}$ & 0.9342 $_{\pm 0.01}$ \\
Four Objs & 0.3202 $_{\pm 0.03}$ & 0.4469 $_{\pm 0.04}$ & 0.6284 $_{\pm 0.02}$ & 0.9400 $_{\pm 0.05}$ & 0.9322 $_{\pm 0.05}$ \\
\bottomrule
\end{tabular}
\label{tb:mlps}

}
\end{table}

\begin{figure}[ht]
    \centering
    \includegraphics[width=\linewidth]{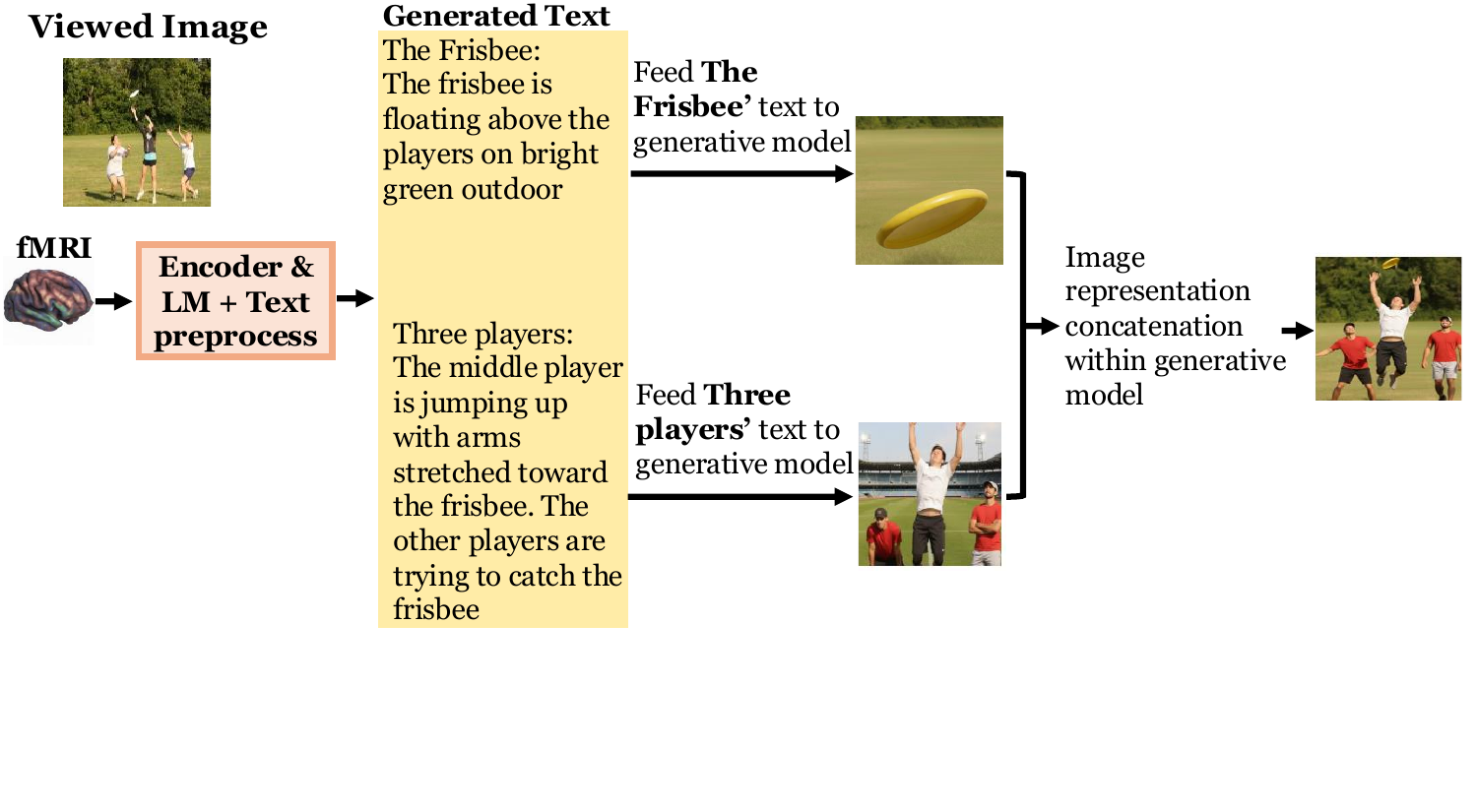}
    \vspace{-2cm}
    \caption{Reconstructed images using two objects.}
    \label{fig:m2_4_succ}
\end{figure}

\begin{figure}[ht]
    \centering
    \includegraphics[width=\linewidth]{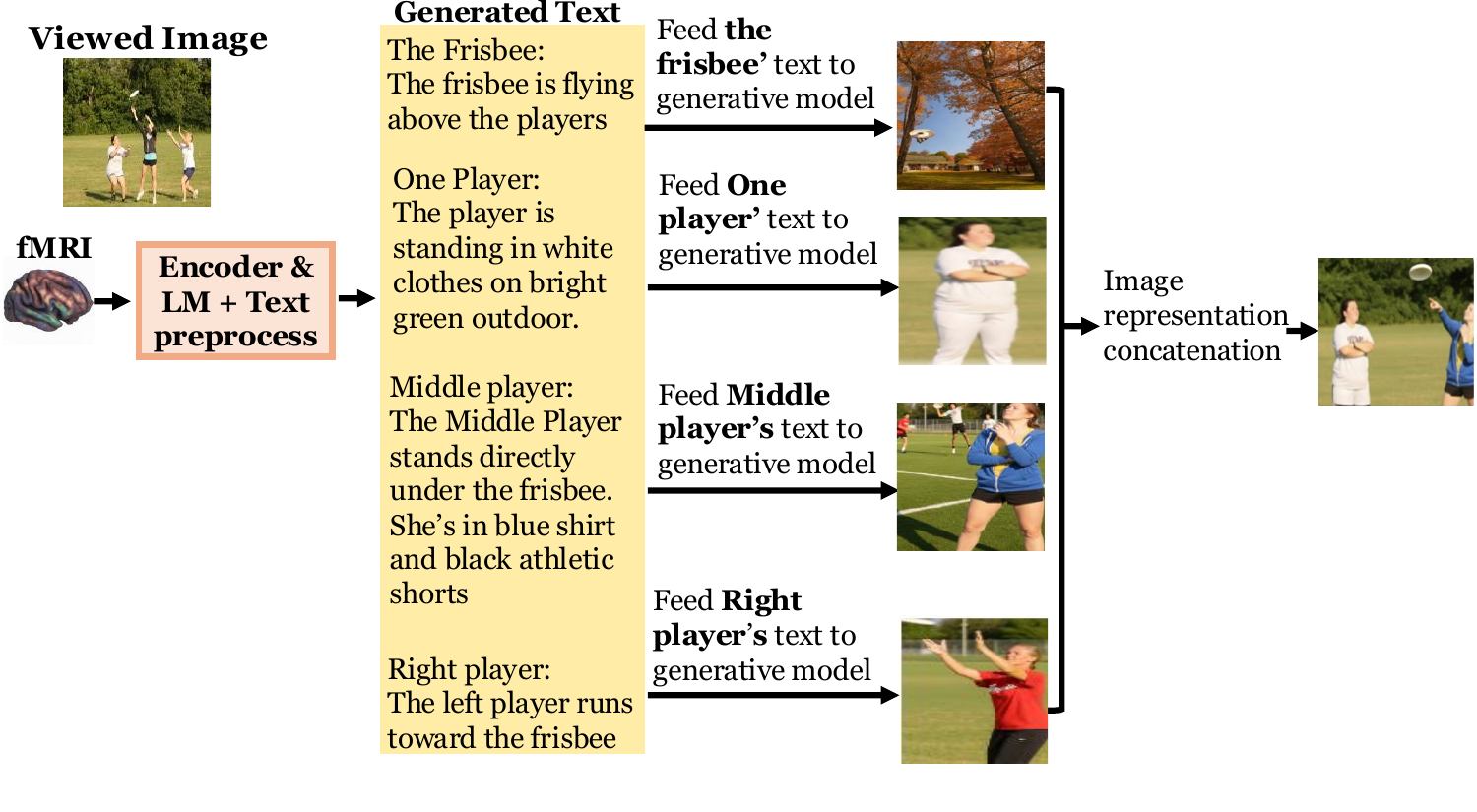}
    \vspace{-1cm}
    \caption{Reconstructed images using four objects.}
    \label{fig:m2_4}
\end{figure}

\textcolor{black}{
We conduct an experiment to determine the optimal number of objects (MLPs) in our framework, and the results on the NSD dataset are reported in Table~\ref{tb:mlps}. The results show that setting the number of objects per image to two yields the best performance. We further present case studies for setting $m=2$ and $m=4$ in Figures~\ref{fig:m2_4_succ} and~\ref{fig:m2_4}, respectively. We observe that for more complex images (i.e., those containing four objects), setting 
$m=2$ encourages the VLM to describe the first object as a group of similar instances, rather than generating isolated descriptions for each object. As shown in Figure~\ref{fig:m2_4_succ}, the individual players are grouped and described as “three players.” In contrast, setting $m=4$ prompts the VLM to generate object-level descriptions (e.g., generating one description per player, as shown in Figure~\ref{fig:m2_4}), leading the generative model to produce distinct latent representations for each object. These representations are then concatenated by spatial location and passed to the pre-trained diffusion model. However, when too many object-level features are introduced, the diffusion model tends to omit objects during the denoising process \cite{liu2024correcting, wang2023imagen}, resulting in incomplete reconstructions—for example, with one of three players missing in the final output (Figure~\ref{fig:m2_4_succ}).
}

% This choice is also supported by neuroscientific findings suggesting that, although the number of objects in an image is inherently uncertain, human attention and memory are limited to only a few of the objects. Both empirical experiments~\cite{alvarez2007many} and neural evidence~\cite{cowan2001magical, todd2004capacity} show that humans can attend to only 3–4 simple objects (e.g., a circle on a white background) at a time. For complex objects (e.g., those with intricate color patterns), this capacity drops to around 2 due to the increased cognitive load per item~\cite{xu2006dissociable}. 

% This cognitive bottleneck limits the amount of information that can be decoded from fMRI signals. As a result, increasing the number of $m$ does not necessarily enhance the level of detail in the reconstructed image and may even lead to less reliable reconstructions—for example, by hallucinating non-existent objects. Therefore, we choose to use $m=2$, as it empirically yields the best performance.

\section{Attribute/relationship Optimization Details}
\label{app:attr_opt}

The detailed algorithm used to solve the prompt optimization problem in~\Cref{eq:opt_obj} is provided in Algorithm~\ref{alg:attr_opt}.
In our experiments, we adopt the scoring function $\mathcal{S}(\mathbf{Y}_1,\mathbf{Y}_2) = 1 - \text{LPIPS}(\mathbf{Y}_1,\mathbf{Y}_2)$. The CKA threshold $\beta$ is initialized to the minimum CKA score among the initial candidates.
We set $\varepsilon=0.5,k_1=8,k_2=2$ and search for $T=40$ rounds. We randomly sampled 667 images from the training set of \textbf{NSD}~\citep{allen2022massive} for prompt optimization. While this subset is used for efficiency, our method is applicable to the full training set and generalizes to other settings. We use \texttt{GPT-4o-mini} as the VLM and the LLM that generates new keywords. We use CLIP-text \citep{radford2021learning} as $\text{LM}_{\text{ENC}}$, and Stable Diffusion 2.1 \citep{pernias2023wurstchen, rombach2022high} as the diffusion model.

\begin{algorithm}[ht]
   \caption{$\varepsilon$-Greedy Prompt Optimization}
   \label{alg:attr_opt}
\begin{algorithmic}
   \STATE \textbf{Input:} Training set $\mathcal{X}^{\text{train}}, \mathcal{Y}^{\text{train}}$; Initial keyword set $\mathcal{A}$; Search rounds $T$; Parameters $\varepsilon, k_1, k_2$
   \STATE \textbf{Initialize:} Threshold $\beta \leftarrow \min_{a \in \mathcal{A}} \text{CKA}\left(\mathbf{X}, \mathbf{K}^a\right)$
   \FOR{$t = 1$ to $T$}
       \STATE \textbf{Filter:} $\mathcal{A} \leftarrow \{a \in \mathcal{A} \mid \text{CKA}\left(\mathbf{X}, \mathbf{K}^a\right) > \beta\}$
       \STATE \textbf{Sort:} Rank $a\in\mathcal{A}$ in descending order by $\sum_{i=1}^N\mathcal{S}\left(\mathbf{Y}_i,\text{Diff}(\text{VLM}(\mathbf{Y}_i,\mathcal{P}(a)))\right)$
       \IF{\texttt{random()} $< \varepsilon$}
           \STATE \textbf{Sample:} $\mathcal{S} \leftarrow \text{RandomSample}(\mathcal{A}, k_1)$ \hfill\% Randomly sample $k_1$ keywords from $\mathcal{A}$
       \ELSE
           \STATE \textbf{Select:} $\mathcal{S} \leftarrow \text{Top}(\mathcal{A}, k_1)$ \hfill\% Select top-$k_1$ keywords from $\mathcal{A}$
       \ENDIF
       \STATE \textbf{Generate:} Use LLM to synthesize $k_2$ new keywords $\mathcal{A}_{\text{new}}$ based on $\mathcal{S}$
       \STATE \textbf{Update:} $\mathcal{A} \leftarrow \mathcal{A} \cup \mathcal{A}_{\text{new}}$
   \ENDFOR
   \STATE \textbf{Output:} $\arg\max_{a \in \mathcal{A}} \sum_{i=1}^N\mathcal{S}\left(\mathbf{Y}_i,\text{Diff}(\text{VLM}(\mathbf{Y}_i,\mathcal{P}(a)))\right)$
\end{algorithmic}
\end{algorithm}

\begin{figure}[H]
    \centering
    \includegraphics[width=\linewidth]{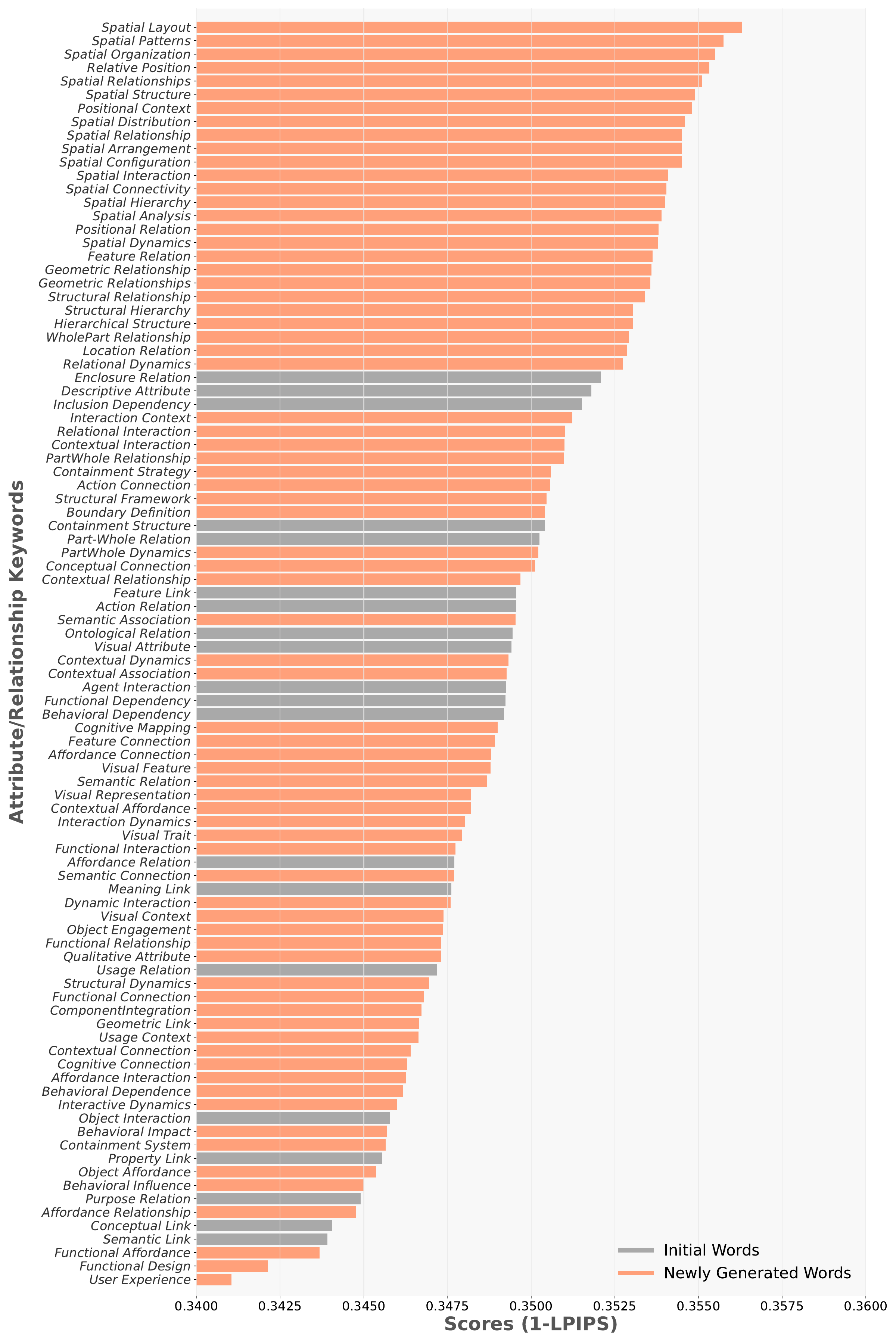}
    \caption{Scores of all keywords during the prompt search. Among the top-10 keywords, the most frequent keyword is 'spatial'.}
    \label{fig:keywords}
\end{figure}

The search is initialized with six widely-used keywords describing object attributes and relationships: Semantic Relationship~\citep{johnson2015image}, Positional Relationship~\citep{lu2016visual,haldekar2017identifying}, Functional Attributes~\citep{zhu2015understanding}, Action Attributes~\citep{lu2016visual}, Visual Attributes~\citep{farhadi2009describing}, and Part–Whole Relationship~\citep{lu2016visual}. For each type, we use \texttt{GPT-4o} to generate four synonymous keywords, resulting in an initial pool of 24 candidate keywords. Figure~\ref{fig:keywords} reports the LPIPS scores of all initial and subsequently discovered keywords.

{\color{black}
\section{Keyword Search with Different Initial Sets}
\label{app:new_keyword_search}
To further demonstrate the effectiveness of our attribute/relationship search module, we conduct an additional experiment where we excluded spatial-related keywords from the initial keyword set. We then evaluated whether our search procedure could still find spatially relevant terms. As shown in Table~\ref{tab:keyword_evolution_app}, the module successfully identifies spatial keywords under this constraint, demonstrating the robustness and effectiveness of our search process.
}

\begin{table}[h]
{
\color{black}
\caption{Top-5 discovered keywords across search rounds.}
\label{tab:keyword_evolution_app}
\setlength{\tabcolsep}{6pt}
\footnotesize
\begin{tabular}{lcccc}
\toprule
\textbf{Rank} & \textbf{Round 0} & \textbf{Round 10} & \textbf{Round 20} & \textbf{Round 30} \\
\midrule
\#1 & Textural Attribute    & Spatial Relationship & Spatial Arrangement & Spatial Arrangement \\
\#2 & Conceptual Linkage    & Textural Attribute   & Proximity Dynamics  & Proximity Relation  \\
\#3 & Action Interaction    & Visual Harmony       & Spatial Relationship & Proximity Dynamics \\
\#4 & Chromatic Feature     & Emotional Resonance  & Color Arrangement   & Spatial Relationship \\
\#5 & Attribute Description & Conceptual Linkage   & Hue Contrast        & Color Arrangement \\
\bottomrule
\end{tabular}
}
\end{table}

\section{Prompts used by \modelname}
% The prompt used by the \texttt{GPT-4o-mini} is shown below:

\section{The Use of Large Language Models (LLMs)
}
We declare that Large Language Models (LLMs) were confined to peripheral tasks and had no influence on the methodology, results interpretation, or theoretical insights of this work. Specifically, they were used for (i) generating training datasets required for our experiments and (ii) grammar correction and minor word-level refinements. All language edits were carefully reviewed by the authors to ensure that no hallucinations were introduced and that the text faithfully reflects the original intent. The technical development, experimental design, analysis, and conclusions presented here are entirely the work of the authors.